\documentclass{article}
\usepackage[preprint]{log_2024}				

\usepackage{booktabs}						
\usepackage{multirow}						
\usepackage{amsfonts}						
\usepackage{graphicx}						
\usepackage{duckuments}						
\usepackage{dsfont}
\usepackage{amssymb}

\newcommand{\xainode}{\textsc{xAI-DropNode}\xspace}
\newcommand{\xaiedge}{\textsc{xAI-DropEdge}\xspace}
\newcommand{\dropmessage}{\textsc{DropMessage}\xspace}
\newcommand{\dropnode}{\textsc{DropNode}\xspace}
\newcommand{\dropedge}{\textsc{DropEdge}\xspace}
\newcommand{\mstd}[2]{\ensuremath{#1{\scriptstyle \pm #2}}}
\usepackage{algorithm}
\usepackage{algpseudocode}
\usepackage{colortbl}

\newcommand{\col}{\cellcolor[HTML]{EFEFEF}}
\newcommand{\xaidrop}{\textsc{xAI-Drop}\xspace}

\newcommand{\al}[1]{#1}

\newcommand{\rev}[1]{{#1}}
\newcommand{\revsecond}[1]{{#1}}
\newcommand{\revthird}[1]{{#1}}

\usepackage[numbers,compress,sort]{natbib}	

\title{xAI-Drop: Don't Use What You Cannot Explain}

\author[]{%
Vincenzo Marco De Luca\\
University of Trento\\
\email{vincenzomarco.deluca@unitn.it}\And
Antonio Longa\\
University of Trento\\
\email{antonio.longa@unitn.it}\And
Pietro Liò\\
University of Cambridge\\
\email{pl219@cam.ac.uk}\And
Andrea Passerini\\
University of Trento\\
\email{andrea.passerini@unitn.it}
}

\begin{document}

\maketitle

\begin{abstract}

Graph Neural Networks (GNNs) have emerged as the predominant paradigm for learning from graph-structured data, offering a wide range of applications from social network analysis to bioinformatics. 
Despite their versatility, GNNs face challenges such as lack of generalization and poor interpretability, which hinder their wider adoption and reliability in critical applications. 
Dropping has emerged as an effective paradigm for improving the generalization capabilities of GNNs. 
However, existing approaches often rely on random or heuristic-based selection criteria, lacking a principled method to identify and exclude nodes that contribute to noise and over-complexity in the model.
In this work, we argue that explainability should be a key indicator of a model's quality throughout its training phase. 
To this end, we introduce xAI-Drop, a novel topological-level dropping regularizer that leverages explainability to pinpoint noisy network elements to be excluded from the GNN propagation mechanism. 
An empirical evaluation on diverse real-world datasets demonstrates that our method outperforms current state-of-the-art dropping approaches in accuracy, and improves explanation quality.

\end{abstract}

\section{Introduction}\label{sec:intro}

The capacity to effectively process networked data has a wide range of potential applications, including recommendation systems~\cite{fan2019graph}, drug design~\cite{Jiang2021}, and urban intelligence~\cite{jiang2022graph}. Graph Neural Networks (GNN)~\cite{1555942,kipf2016semi,gilmer2020message,gnnsurvey2021} have emerged as a powerful and versatile paradigm to address multiple tasks involving networked data, from node and graph classification~\cite{gilmer2017neural,ying2018graph,tang2022graph,bianchi2024expressive} to link prediction~\cite{li2022collaborative,zhang2018link,lachi2024a} and graph generation~\cite{you2018graphrnn,vignac2022digress,jaeger2024simple}.

Despite their effectiveness and popularity, GNNs face various challenges that prevent their wider adoption and reliability in critical applications~\cite{waikhom2021graph,liao2021review,liang2022survey}, such as lack of generalization~\cite{garg2020generalization}poor interpretability~\cite{longa2022explaining}, \rev{oversmoothing~\cite{xuanyuan2023shedding, rusch2023survey}, and oversquashing~\cite{giraldo2023trade}}. A significant challenge for GNNs is their vulnerability to noise, as irrelevant or noisy node features can propagate through the layers and degrade the model’s performance. Dropping~\cite{li2018deeper} has emerged as an effective paradigm to reduce noise and improve GNN robustness. Dropping can be performed at different granularities, from dropping single messages~\cite{fang2023dropmessage} to dropping edges~\cite{rong2019dropedge}, or even nodes~\cite{fan2019graph,papp2021dropgnn}. However, existing approaches often rely on random or heuristic-based selection criteria, and lack a principled method to identify and exclude nodes that contribute to noise and over-complexity in the model.



\begin{figure}[t!]
    \centering
    \includegraphics[width=\linewidth]{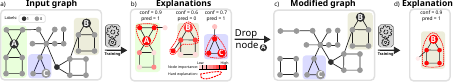}
    \caption{\rev{Illustration of the rationale behind \xaidrop. Panel (a) shows a Barabási-Albert network with house-shaped motifs randomly attached. The task here is to classify nodes as either the top of a house (label 1) or otherwise (label 0). It is easy to see that a triangle is an approximate pattern for the positive class. The figure highlights three prototypical nodes (A, B, C) which are parts of a triangle, where only two (A, B) are also the top of a house (triangle and houses highlighted for readability). Panel (b) reports the explanation of a GNN trained on the network, for the three highlighted nodes. Node A has a high confidence, because it has both the correct (the house) and spurious (the two triangles) patterns. However its explanation is mostly based on the (simpler) spurious triangle, which is insufficient to explain its confidence (as shown by the lower confidence of nodes B and C). Removing A (Panel c) prevents the network from focusing on the spurious pattern, so that the correct pattern is eventually learned (Panel d, with C omitted as no longer predicted as label 1).}}
    \label{fig:mainfig}
\end{figure}

In this paper we argue that explainability should be considered a first class citizen in determining which elements of the graph should be dropped in order to increase the robustness of the learned GNN. Consider a GNN being trained for node classification. Our intuition is that the fact that the prediction for a given node has a poor explanation is a symptom of a suboptimal function being learned, and that this symptom is more harmful if the prediction has a high-confidence. \rev{Figure~\ref{fig:mainfig} provides a graphical illustration of this intuition.} Guided by this rationale we present \xaidrop, a novel topological-level dropping regularizer that leverages explainability and over-confidence to pinpoint noisy network elements to be excluded from the GNN propagation mechanism during each training epoch. 

An empirical evaluation on diverse real-world datasets demonstrates that our method outperforms current state-of-the-art dropping approaches in accuracy, and improves explanation quality.
Our main contributions can be summarized as follows:

\begin{itemize}

\item We identify local explainability {\em during training} as a driving principle to discard noisy information in the GNN learning process.

\item We introduce \xaidrop, an explainability-guided dropping framework for GNN training.

\item We show how \xaidrop consistently outperforms alternative dropping strategies as well as XAI-based regularization approaches on various node classification and link prediction benchmarks.

\item We demonstrate the effectiveness of \xaidrop in improving explanation quality.

\end{itemize}

The rest of the paper is organized as follows. We start by reviewing related work (Section~\ref{sec:related}) and then introduce the relevant background (Section~\ref{sec:prelim}). Our \xaidrop framework is presented in Section~\ref{sec:xaidrop} and experimentally evaluated in Section~\ref{sec:exps}. Finally, conclusions are drawn in Section~\ref{sec:concl}.


\section{Related Work}
\label{sec:related}

\textbf{Dropping.} 
Dropping strategies are commonly used in neural networks to prevent overfitting~\cite{srivastava2014dropout} by randomly setting a portion of neurons to zero during training, which helps the network learn more robust features. In GNNs, this approach has been extended to the topological level, altering message propagation between nodes\rev{, often to reduce oversmoothing~\cite{xuanyuan2023shedding}}. DropEdge~\cite{rong2019dropedge} was the first to introduce this concept by randomly dropping edges during training based on a Bernoulli distribution. Inspired by DropEdge, subsequent methods include DropNode~\cite{feng2020graph} which drops nodes and their connections, and its variants DropGNN~\cite{papp2021dropgnn}, which removes nodes also at test time; DropMessage~\cite{fang2023dropmessage}, which drops messages during propagation;  and DropAGG~\cite{jiang2023dropagg}, which randomly selects neighboring nodes for aggregation. While these methods use random sampling, alternative strategies for component dropping have also been explored in the literature.
FairDrop~\cite{spinelli2021fairdrop}  combines randomicity and fairness to adjust graph topology for link prediction tasks. Learn2Drop~\cite{luo2021learning} is a learnable graph sparsification procedure deciding which edges to drop in order to retain maximal similarity to the original network. Beta-Bernoulli Graph Drop Connect (BBGDC) ~\cite{hasanzadeh2020bayesian} adapts the drop rate of the edges during training based on a Beta-Bernoulli distribution. All these methods 
rely on random or heuristic-based selection criteria. In this work we show how a more principled XAI-based method to identify potentially harmful components substantially outperforms existing dropping strategies.









\textbf{Post-hoc explanability.} Several works investigate post-hoc methods to explain the predictions of GNN models. GNN explainers can be categorized into model-level and instance-level explainers. Model-level explainers~\cite{wang2023gnninterpreter,azzolin2023global,yuan2020xgnn} aim at providing a global understanding of a trained model, e.g., as motifs or rules driving the model to predict a certain class. In contrast, instance-level explainers~\cite{ying2019gnnexplainer,vu2020pgm,miao2022interpretable,yuan2021explainability} aim at identifying components of a given input that are responsible for the model’s prediction for that input, and are thus more appropriate to design an XAI-based dropping strategy. Instance-level explainers can be grouped into five categories~\cite{kakkad2023survey}: decomposition, surrogate, gradient, perturbation and generation based. Decomposition-based methods break down the input to identify explanations~\cite{pope2019explainability}. Surrogate-based methods rely on an interpretable surrogate to explain the prediction of the original model~\cite{huang2022graphlime,vu2020pgm}. Gradient-based methods define explanations in terms of the gradient of the network output with respect to the elements of the input graph~\cite{sundararajan2017axiomatic,pope2019explainability}. Perturbation-based methods manipulate the input to obtain interpretable subgraphs~\cite{ying2019gnnexplainer,luo2020parameterized}, while generation-based methods generate subgraphs that can explain the model output~\cite{li2022dag}. 
\rev{In this work, we used a gradient-based method, specifically an approximation of the saliency map~\cite{simonyan2013deep}, due to its computational efficiency (see Appendix~\ref{app:aproxSal}). However, our framework is flexible and can be applied to any explainer that produces node-level explainability scores (see Appendix~\ref{app:other_explainers}).}

\textbf{XAI-based regularization.} \revthird{A number of xAI-driven approaches have been proposed to enhance the performance of deep learning methods~\cite{weber2023beyond}, from addressing interactive data augmentation~\cite{schramowski2020making} to enabling automated pruning~\cite{yeom2021pruning}.} A few approaches have been recently proposed to explicitly introduce XAI-based regularization strategies during the training stage of GNNs. MATE~\cite{spinelli2022meta} applies an optimization procedure via meta-learning to enhance explainability of the resulting model. ExPass~\cite{giunchiglia2022towards} works at the message passing level by weighting messages with the importance of nodes as defined by PGExplainer~\cite{luo2020parameterized}, while ENGAGE~\cite{shi2023ENGAGE} presents Smoothed Activation Maps~\cite{shi2023ENGAGE} to perturb low scores edges and features. These methods however fail to consider the quality of the explanation and are heavily parameterized, resulting in substantial computational overhead, learnability issues, and eventually suboptimal performance. 
Our experimental evaluation shows how our simple XAI-driven dropping strategies outperform these methods in terms of {\em both} accuracy and explainability.

\section{Preliminaries}
\label{sec:prelim}
In this section, we provide an overview of the fundamental concepts underlying our approach.


\paragraph{Graph.} A graph is a tuple $G = (\mathcal{V}, \mathcal{E}, X_\mathcal{V}, X_\mathcal{E})$, where $\mathcal{V}$ is a set of vertices or nodes, $\mathcal{E}$ is a set of edges between the nodes,  $X_\mathcal{V}$ and $X_\mathcal{E}$ are node features and edge features, respectively. Node and edge features may be empty. The set of edges $\mathcal{E}$  can be represented as an adjacent matrix $A \in R^{|\mathcal{V}|\times |\mathcal{V}|}$, where $A_{ij} = 1$ if $(v_i,v_j) \in \mathcal{E}$, 0 otherwise. In this paper we will focus on undirected graphs, in which edges have no directions, i.e., $A_{ij} = A_{ji}$. Given $v \in \mathcal{V}$, the set $\mathcal{N}_v = \{ u \in \mathcal{V}: (u,v) \in \mathcal{E} \}$ denotes the neighborhood of $v$ in $G$.

\paragraph{Graph Neural Network (GNN)} A GNN is a class of neural network architecture specifically designed to process graph data~\cite{4700287,micheli2009neural,kipf2017semisupervised}. A GNN leverages a message-passing scheme to propagate information across nodes in a graph. GNNs iteratively learn node representations $\mathbf{h}_{v}$ by aggregating information from neighboring nodes. 
%
%
%
In most cases, the propagation mechanism for an entire layer can be compactly represented using the adjacency matrix $A$, the node embedding matrix $H^{(l)}$ and one or more layer-specific weight matrices $ W^{(l-1)}$. For instance, layerwise propagation in GCN~\cite{kipf2017semisupervised} can be written as:
$$
H^{(l)} = \sigma\left(\tilde{D}^{-\frac{1}{2}} \tilde{A} \tilde{D}^{-\frac{1}{2}} H^{(l-1)} W^{(l-1)} \right)
$$
where $\tilde{A} = A + I_{|\mathcal{V}|}$ is the adjacency matrix enriched with self loops, $\tilde{D}_{ii} = \sum_{j} \tilde{A}_{ij}$ and $\sigma$ is a non-linear activation function such \text{ReLU} or sigmoid. Dropping strategies, including our \xaidrop approach, can be formalized in terms of modifications to the adjacency matrix $A$.

\paragraph{GNN explainability.} 
Intuitively, given a graph $G$ and a trained GNN $f$, an explanation is a subgraph $G_{exp} \subset G$ that contains the information that is relevant for $f$ to perform inference on $G$. We use $G_{exp}(v)$ to denote the local explanation for the GNN output for node $v$. 
In this study, we employ the saliency map method \cite{simonyan2013deep}, an instance-based and gradient-based explainer that computes the attribution for each input by performing backpropagation to the input space. The general idea is that the magnitude of the derivative provides insights into the most influential features, which, when perturbed, result in the highest difference in the output space. Formally, it is defined as
\begin{equation}\label{eq:sal}
    G_{exp}(v) =  \frac{\partial f_v(G)}{\partial X_v}
\end{equation}
Where $f_v(G)$ is the prediction of the model for node $v$, and $X_v$ is the feature vector of node $v$.

\textbf{Fidelity sufficiency ($F_{suf}$).} Fidelity sufficiency~\cite{rathee2022bagel} is a popular explainability metric for GNNs. It measures the distance between the probability predicted by $f$ when fed with the entire graph $G$ and the probability when fed with the explanation $G_{exp}$ respectively:

\begin{equation}\label{eq:suf}
    F_{suf}(v) = 1 - d(f_v(G), f_v(G_{exp}(v)))
\end{equation}

with $d(\textbf{p},\textbf{p}')$ being a distance over probability distributions, which in our work we have identified with the Kullback–Leibler divergence.




\section{Explainability-Based Dropping}
\label{sec:xaidrop}

\begin{figure*}[h!]
    \centering
    \includegraphics[width=0.9\linewidth]{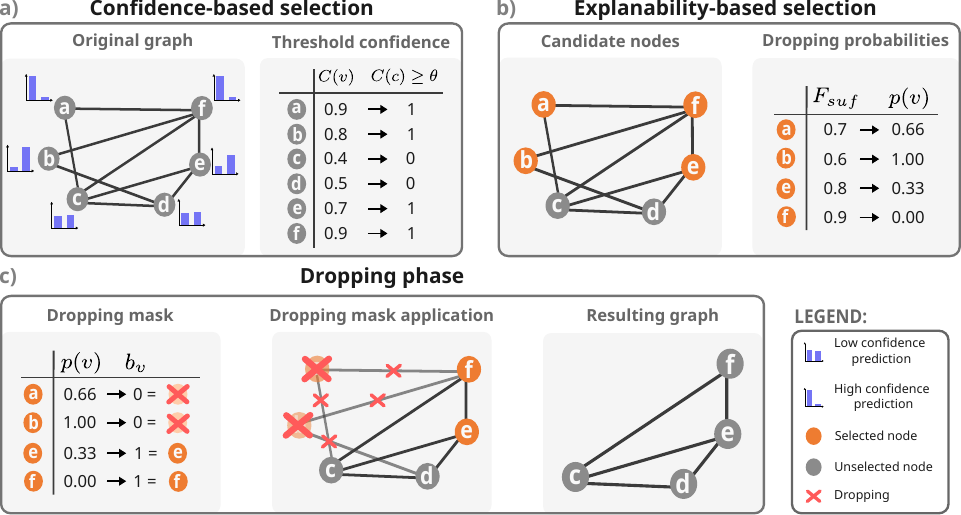}
    \caption{A graphical representation of the node dropping strategy (\xainode) employed by the \xaidrop algorithm in node classification tasks. Panel \textbf{a)} illustrates the confidence-based selection process, where nodes are selected if the model’s confidence is equal or greater than a specified threshold $\theta$. Panel \textbf{b)} presents the computation of fidelity sufficiency scores and dropping probabilities for the nodes selected in panel \textbf{a)}. Lastly, panel \textbf{c)} illustrates the computation of the node dropping mask by Bernoulli sampling, and the resulting graph after dropping nodes and their associated edges.}
    \label{fig:pipeline}
\end{figure*}

\xaidrop is based on the combination of two concepts: explainability and (over)confidence. On the one hand, a poor local explanation can be seen as a symptom of an unreliable prediction for the corresponding node, making it a good candidate for being dropped to reduce noise during training. On the other hand, a highly confident prediction for a node indicates that the network is very confident about the features the prediction is based upon, that in principle should correspond to the local explanation. A confident prediction with a poor explanation is thus a combination one would like to avoid as much as possible. Building on these intuitions, 
\xaidrop is a general framework that implements a dropping strategy that targets samples with poor explanations and high certainty. 
In presenting the \xaidrop framework we focus on the node classification task (and node dropping) with \xainode, but the method can be readily applied to link prediction, as discussed in Section~\ref{sec:dropedge}. In the following we focus on the transductive setting, which is by far the most common in node classification. The approach can also be applied to inductive settings, like graph classification tasks, as discussed at the end of the section. 

Figure~\ref{fig:pipeline} presents a graphical representation of the \xainode approach, which consists of two main phases: node selection and dropping. The node selection phase (further detailed in Section \ref{sub:node_sel}) consists of four steps.
In the first step, the most certain nodes are extracted as candidates for dropping, by setting a threshold $\theta$ ($\theta=0.7$ in Figure~\ref{fig:pipeline}) over the predicted confidence for the most probable class.
In the second step, the fidelity sufficiency score of these nodes is computed using Eq.~\ref{eq:suf} to assess the local explanation of these predictions. This score is then mapped into a dropping probability $p(v)$ for the node by applying an appropriate transformation (detailed in Section~\ref{sub:node_sel}), such that the nodes having the worst explanations will have the highest probability of being dropped. Finally, a decision on whether to retain or drop each candidate node $v$ is made according to a Bernoulli distribution parameterized by $p(v)$.

\subsection{Node selection}\label{sub:node_sel}
First, a forward step is computed on the entire set of nodes $\mathcal{V}$, including training, validation, and test nodes. The goal of this first step is to obtain the confidence score for each node $v$ which is computed as: 
\begin{equation}\label{eq:conf}
    C(v) = \text{max}_y P(y|X_v)
\end{equation}
where $X_v$ is the feature vector associated with node $v$.
From this large set of nodes, only the most confident nodes are selected as the candidate dropping set, $\mathcal{V}' \subset \mathcal{V}$. For each node $v \in \mathcal{V}'$, its local explanation $G_{exp}(v)$ is computed using an \al{approximation of the saliency map~\cite{simonyan2013deep} as explainer, which helps to further reduce the computational overhead of generating explanations (see Appendix \ref{app:aproxSal} and \ref{app:time} for details).}
We opted for a gradient-based explainer because it is not computationally demanding and do not require ground truth explanations, but the method is agnostic with respect to the explanation method being used. Fidelity sufficiency scores are then computed according to Eq.~\ref{eq:suf}. Note that the metric requires a hard explanation, while the produced explanations are soft masks  (i.e., a real value associated with each node indicating its importance for the prediction being explained). In this manuscript, we discretize soft explanations by selecting the top 25\% of the edges as part of the hard explanation. \revsecond{Nonetheless, the approach can in principle be applied to soft explanations by weighting messages according to the generated explanations.}\\



The next step consists in assigning dropping probabilities to the nodes in $\mathcal{V}'$, such that worse explanations, i.e., low fidelity sufficiency, correspond to higher dropping probabilities. Given a predefined dropping probability $p$, the idea is to adjust dropping probabilities for individual nodes according to their fidelity sufficiency, without affecting the expected fraction of nodes to be selected for dropping. 
The dropping probability of node $v \in \mathcal{V}'$ is adjusted by mapping the fidelity sufficiency scores into a Gaussian distribution by applying the Yeo-Johnson transformation~\cite{yeo2000new}
which is defined as follows:
\begin{equation}\label{eq:yeo-johnson}
    \phi(x; \lambda) = \begin{cases}
      \frac{(x+1)^{\lambda}-1}{\lambda} & \text{ }  \lambda \neq 0, x \geq 0\\
      log(x+1) & \text{ }  \lambda = 0, x \geq 0 \\
      -\frac{(-x+1)^{2-\lambda}-1}{2-\lambda} & \text{ }  \lambda \neq 2, x < 0\\
      -log(-x+1) & \text{ }  \lambda=2, x < 0
    \end{cases}\
\end{equation}
Where $x$ is the response variable, which is set to  $x = 1 - F_{suf}$ in our case (lower sufficiency implies higher dropping probability), and $\lambda$ is a learnable parameter. The parameter is selected via log-likelihood maximization so as to maximize the normality of the transformed data~\cite{yeo2000new}. Values are then normalized and shifted so as to have mean equal to $p$. \rev{Preliminary experiments showed that this solution achieves better results with respect to using the empirical distribution }\revsecond{ (Appendix \ref{app:exp_mapping_prob}.)}

All nodes $u \in \mathcal{V} \setminus \mathcal{V}'$ that do not belong to the confidence node subset retain the default dropping probability, $p(u) = p$. Overall, this procedure \rev{biases the dropping probability to encourage the dropping of potentially noisy nodes, while guaranteeing} that the expected number of nodes selected for dropping is equal to the predefined dropping probability $p$. \rev{An empirical evaluation of different strategies to detect candidate noisy nodes is reported in Appendix~\ref{app:candidate_noisy_appendix}}.

\subsection{Dropping}\label{sub:drop}
Once the biased dropping probabilities $p(v)$ have been computed, they can be employed to alter the propagation of information to regularize the learning. In detail, \xainode removes nodes  
from the node set $\mathcal{V}$ based on a node dropping mask $\mathbf{b} \in \{0,1\} ^ {|\mathcal{V}|}$ defined as follows:
\begin{equation}\label{eq:xai-drop_node}
b_v  \sim Bernoulli(1-p(v))
\end{equation}
Once a node is dropped, all its incident edges $\mathcal{I}_v = \{ (u,w) \in \mathcal{E}: u = v \textrm{ or } w = v \}$ are also removed. Following~\cite{fang2023dropmessage}, this operation can be compactly represented in terms of a modified adjacency matrix:
\begin{equation}
\label{eq:dropnode_adj}
A' = B \; A \;  B
\end{equation}
\noindent where $B$ is a diagonal matrix having the elements of $\mathbf{b}$ on the main diagonal (and zero elsewhere).

\subsection{Overall procedure}
The overall algorithm for \xaidrop is outlined in Algorithm \ref{algo:algo1}. The algorithm takes as input a graph $G$, the GNN architecture to be trained $f$, and the hyper-parameters $\theta$ and $p$,\rev{ for further detail about these hyper-parameters refer to Appendix~\ref{hyperparameter_and_sensitivity}}.
In each epoch, the algorithm selects the nodes with a prediction confidence score of at least $\theta$, and computes their explainability according to the current version of $f$ in terms of fidelity sufficiency. Fidelity sufficiency values (collectively indicated as $F_{suf}$) are then used to determine the node-specific dropping probabilities, guaranteeing that a fraction $p$ of the nodes are being dropped in expectation. These probabilities are in turn used to select nodes (and corresponding incident edges) to be dropped and produce an adjusted adjacency matrix $A'$. Finally, $A'$ is used for another round of training of the GNN $f$. Note that the dropping procedure is only performed at training time. Indeed, the inference procedure (last line of the algorithm) employs the full adjacency matrix $A$, consistently with what is done by existing dropping strategies in the literature.


\begin{algorithm}
\caption{\xaidrop algorithm for node classification tasks. $G=(\mathcal{V},\mathcal{E},\textbf{X}_\mathcal{V},\textbf{X}_\mathcal{E})$ is a graph, $f$ is the GNN, $\theta, p$ are hyper-parameters}
\label{algo:algo1}
\begin{algorithmic}[1]
\Procedure{\xaidrop}{$G=(\mathcal{V},\mathcal{E},\textbf{X}_\mathcal{V},\textbf{X}_\mathcal{E})$,$f$, $\theta$, $p$}
    \For{$e \in \text{Epochs}$}
        \State $\mathcal{V}' \leftarrow \textsc{highest-confidence}(G,\mathcal{V},f, \theta$) \Comment{Equation \ref{eq:conf}} 
        \For{$v \in \mathcal{V}'$}
            \State $G_{exp}(v) \leftarrow \textsc{saliency-map}(G,v)$  \Comment{Equation \ref{eq:sal}}
             \State $F_{suf}(v) \gets \textsc{fidelity}(f,G,G_{exp}(v))$ \Comment{Equation \ref{eq:suf}}
        \EndFor
        \State $\mathbf{p} \leftarrow \textsc{dropping-probabilities}(F_{suf}, p)$
        \Comment{Equation \ref{eq:xai-drop_node}}
        \State $A' \rightarrow \xainode(G,\mathbf{p})$ \Comment{Equation \ref{eq:dropnode_adj}}
        \State $f \gets \textsc{train}(f, G, A')$
    \EndFor
    \State $\mathcal{Y} \gets \textsc{evaluate}(f, G, A)$
\EndProcedure
\end{algorithmic}
\end{algorithm}

\subsection{\xaidrop for Link Prediction}
\label{sec:dropedge}

While we focused on node classification in describing the approach for the sake of clarity, the \xaidrop framework can also be applied to link prediction, where the goal is dropping edges which have highly confident but poorly explained predictions. In this setting, that we name \xaiedge, the confidence score is computed at the edge level, rather than the node level, and the explainer will produce an explanation for each edge. The explanations of the edge predictions are assessed by aggregating the scores such that each edge has a corresponding normalized fidelity sufficiency. \xaiedge produces a dropping probability score for each edge in the input graph and then the procedure drops the edges according to a Bernoulli distribution, i.e., $B_{ij}  \sim  Bernoulli(1-p(e_{ij}))$. \rev{See Appendix~\ref{app:dropedge} for further details}. In our experimental analysis we will show the effectiveness of \xaidrop with its two main variants: \xainode for node classification tasks and \xaiedge for link prediction tasks.

\rev{We presented \xaidrop for transductive learning settings, which include unsupervised nodes in the message propagation process. The approach can also be applied to inductive settings, like graph classification, where it boils down to dropping entire training instances.}



\section{Experiments}
\label{sec:exps}

Our experimental evaluation aims to address the following research questions:
\begin{itemize}
    \item[\textbf{Q1:}] Does \xaidrop outperform alternative dropping strategies?
    \item [\textbf{Q2:}] Does \xaidrop outperform alternative xAI-driven strategies?
    \item [\textbf{Q3:}] Does \xaidrop improve explainability?
    \item [\textbf{Q4:}] Does \xaidrop work on beyond node classification? 
\end{itemize}

We start by presenting the experimental setting and then discuss the results answering these questions.

\subsection{Experimental setting}
\label{sec:exp_setting}

\paragraph{\textit{Datasets:}} We employed three widely used datasets for node classification:   Cora~\cite{mccallum2000automating}, Citeseer~\cite{giles1998citeseer}, and PubMed~\cite{yang2016revisiting}. Each dataset is composed of a single graph with thousands of labeled nodes. We utilize the publicly available train, validation, and test node splits~\cite{yang2016revisiting}. We employed the same datasets for link prediction, where we used 10\% of the edges for the validation set, and 20\% of the edges for the test set.  Detailed dataset statistics are presented in appendix \ref{app:data}.

\paragraph{\textit{Competitors:}} We compare our \xaidrop approach with state-of-the-art dropping strategies. Random dropping methods include: \dropmessage~\cite{fang2023dropmessage}, which removes random features from messages; \dropedge~\cite{rong2019dropedge}, which randomly removes edges; \dropnode~\cite{fan2019graph,papp2021dropgnn}, which removes random nodes and all of their incident edges; and DropAgg~\cite{jiang2023dropagg}, which discards messages to sampled nodes during aggregation. Non-random methods include Learn2Drop~\cite{luo2021learning}, using parameterized networks to prune irrelevant edges, and BBGDC~\cite{hasanzadeh2020bayesian}, which adapts edge drop rates during training. As our method is XAI-based, we also compare it to XAI-based GNN regularization techniques: ExPass~\cite{giunchiglia2022towards} adjusts message weights based on importance scores; MATE~\cite{spinelli2022meta} uses meta-learning to optimize explainer performance; and ENGAGE~\cite{shi2023ENGAGE} introduces a novel explainer for data augmentation to enhance GNN robustness.

\paragraph{\textit{GNN Architectures:}} By operating on the adjacency matrix, \xaidrop is  agnostic about the underlying GNN architecture. To demonstrate its versatility, we implemented it with three widely recognized GNN architectures: Graph Convolutional Networks (GCN)~\cite{kipf2016semi}, Graph Attention Networks (GAT)~\cite{veličković2018graph} and Graph Isomorphism Network (GIN)~\cite{xu2018powerful}.
For the random strategies, we retained the hyper-parameters that were optimized in the original work evaluating them~\cite{fang2023dropmessage}. The same holds for ExPass~\cite{giunchiglia2022towards}. For ENGAGE~\cite{shi2023ENGAGE} and MATE~\cite{spinelli2022meta} we optimized hyper-parameters ourselves over the validation set, as the available configuration was not usable (ENGAGE was trained on a larger training set with respect to the standard split, while MATE was mostly evaluated on synthetic data). 
In the case of \xaidrop, we maintained the same GNN hyper-parameters as those used in random strategies to isolate the role of the explainability component. Concerning \xaidrop-specific hyper-parameters, we fixed $p=0.5$ in all settings for simplicity, which implies dropping on average 50\% of the edges/nodes (while the dropping probability $p$ is optimized for each single dataset and GNN architecture in the case of the random strategies~\cite{fang2023dropmessage}), and set the confidence threshold $\theta$ to 0.9, 0.55, and 0.95 for, respectively, Cora, CiteSeer and PubMed,
after a coarse-grained optimization on the validation set.

\paragraph{\textit{Metrics:}}

Multiclass accuracy is used for addressing research questions \textbf{Q1} and \textbf{Q2}. 
To answer research question \textbf{Q3}, we exploited the traditional version of saliency map as explainer as defined in Eq.\ref{eq:sal}, not its approximated variant used for training that we have defined in \ref{app:aproxSal}, and computed the accuracy sufficiency, defined as:

\begin{equation}\label{eq:suf_norm_acc}
    A_{suf}(G) = \frac{1}{|\mathcal{V}_{\text{test}}|}\sum_{v \in \mathcal{V}_{\text{test}}} \mathds{1}\left(\text{argmax}(f_v(G)) = \text{argmax}(f_v(G_{exp}(v)))\right) 
\end{equation}
where $\mathcal{V}_{\text{test}}$ is the set of test nodes in $G$, $G_{exp}(v)$ is the (thresholded version of the) saliency map defined in Eq.~\ref{eq:sal}, and $\mathds{1}(\cdot)$ is the indicator function. While in the case of the link prediction task (\textbf{Q4}) we used Area Under the Curve (AUC) for assessing the model predictions.



\subsection{Experimental results}
\paragraph{{\bf R1: \xaidrop outperforms alternative dropping strategies.}} 

Table~\ref{tab:acc} shows the test accuracy of the different approaches we tested on node classification. Mean and standard deviation over 5 runs with different initialization seeds are reported. Comparing our \xaidrop strategies with the blocks of random and learning-based dropping approaches, the advantage of XAI-driven dropping is evident\footnote{Notice that the results in the table are different from those in the original publications of each respective method, as we had to rerun them all (retaining their optimal hyper-parameters or optimizing them on the validation set, as explained in Section~\ref{sec:exp_setting}) in order to compute explainability metrics in addition to accuracy. Accuracy comparisons with the results reported in the original papers are reported in Appendix~\ref{app:literature_res}, and confirm the advantage of the \xaidrop strategy.}. \xainode consistently outperforms its random counterpart (DropNode which in turn improves over the baseline method with no dropping) on all datasets and for all GNN architectures, despite the fact that the biased dropping probability (Eq.~\ref{eq:xai-drop_node}) is applied to the most confident nodes. More importantly, it outperforms all alternative dropping strategies, both random-based and learning-based\footnote{We omit the results of BBGDC on GAT and GIN, because the method was specifically designed for GCN architectures and it failed to learn usable models when applied to GAT and GIN architectures.}. Indeed, \xainode consistently scores as the best method in almost all scenarios. These results support our intuition that explainability can be an effective metric to guide the identification and removal of noisy information in GNN training.  

\paragraph{{\bf R2: \xaidrop outperforms alternative xAI-driven strategies}} \xaidrop is not the first method to propose the use of explainability to enhance training. The XAI-based block in Table~\ref{tab:acc} reports test accuracy of existing alternative xAI-based regularization methods. These methods underperform with respect to our \xaidrop strategies, likely because of the increased complexity of their training process with respect to our dropping schemes. It is important to remind here that these xAI-based competitors have been developed with additional goals in mind with respect to regularization, namely improving explainability (MATE), alleviating oversmoothing (ExPass), or increasing robustness to adversarial attacks (ENGAGE). For large-scale datasets refer to Appendix~\ref{app:scale-xai-drop}

\begin{table}[h!]
    \centering
    \resizebox{\textwidth}{!}{ 
        \begin{tabular}{l@{\hskip 9pt}l@{\hskip 5pt}|c@{\hskip 6pt}c@{\hskip 6pt}c@{\hskip 6pt}|c@{\hskip 6pt}c@{\hskip 6pt}c@{\hskip 6pt}|c@{\hskip 6pt}c@{\hskip 6pt}c}
        &        & \multicolumn{3}{c|}{\textbf{GCN}} & \multicolumn{3}{c}{\textbf{GAT}} & \multicolumn{3}{c}{\textbf{GIN}}\\
        &Model   & Cora & CiteS & PubM & Cora & CiteS & PubM & Cora & CiteS & PubM \\
        \toprule
            &Baseline & 
                \mstd{79.0}{0.3} & \mstd{67.1}{0.5} & \mstd{76.9}{1.2}& 
                \mstd{78.4}{1.2} & \mstd{68.1}{0.7} & \mstd{77.3}{0.7}& 
                \mstd{78.2}{1.0} & \mstd{67.5}{1.0} & \mstd{76.7}{0.8}\\
        \midrule
        \multirow{4}{*}{\rotatebox[origin=c]{90}{Random}} 
            & DropEdge & 
                \mstd{80.0}{0.5} & \mstd{68.4}{0.6} & \mstd{77.5}{0.4}& 
                \mstd{79.8}{0.3} & \mstd{68.3}{0.7} & \mstd{77.3}{0.4}& 
                \mstd{79.2}{0.7} & \mstd{69.0}{0.8} & \mstd{77.3}{0.6}\\
            &\col{DropMess} & 
                \col{\mstd{80.8}{0.5}} & \col{\mstd{70.8}{0.5}}& \col{\mstd{78.1}{0.3}} &
                \col{\mstd{80.1}{0.6}} & \col{\mstd{69.5}{0.8}}& \col{\mstd{77.5}{0.5}} &
                \col{\mstd{78.8}{0.5}} & \col{\mstd{69.7}{0.5}}& \col{\mstd{77.9}{0.6}} \\
            & DropNode & 
                \mstd{80.0}{0.5} & \mstd{69.4}{0.4} & \mstd{78.0}{0.4}& 
                \mstd{78.6}{1.3} & \mstd{67.5}{0.7} & \mstd{77.4}{0.2}& 
                \mstd{79.4}{1.0} & \mstd{69.4}{0.6} & \mstd{77.2}{0.5}\\
            &\col{DropAggr} & 
                \col{\mstd{80.0}{0.6}} & \col{\mstd{68.8}{0.6}} & \col{\mstd{78.3}{0.3}} &
                \col{\mstd{80.8}{0.8}} & \col{\mstd{67.3}{1.2}} & \col{\mstd{77.7}{0.2}}&
                \col{\mstd{78.6}{0.5}} & \col{\mstd{68.2}{1.6}} & \col{\mstd{76.9}{0.4}} \\ 
        \midrule
        \multirow{2}{*}{{\rotatebox[origin=c]{90}{Learn}}} 
                & BBGDC & 
                \mstd{74.2}{0.3} & \mstd{67.4}{0.3} & \mstd{74.1}{0.6}& - & - & - & - & - & - \\
            & \col{Learn2Drop} & 
                \col{\mstd{79.3}{1.1}} & \col{\mstd{68.6}{0.9}} & \col{\mstd{77.1}{1.4}} &
                \col{\mstd{80.5}{0.7}} & \col{\mstd{70.5}{0.9}} & \col{\mstd{77.4}{0.6}}  &
                \col{\mstd{78.4}{1.4}} & \col{\mstd{68.1}{0.9}} & \col{\mstd{77.0}{1.1}} \\  
        \midrule
        \multirow{3}{*}{\rotatebox[origin=c]{90}{xAI}} & 
            MATE & 
                \mstd{80.3}{0.4} & \mstd{68.4}{0.3} & \mstd{74.3}{0.5}& 
                \mstd{80.0}{0.8} & \mstd{69.2}{0.6} & \mstd{76.2}{0.7}& 
                \mstd{81.1}{0.9} & \mstd{71.2}{1.3} & \mstd{78.8}{1.2}\\     
            & \col{ExPass} & 
                \col{\mstd{82.2}{0.6}} & \col{\mstd{72.9}{0.4}} & \col{\mstd{76.2}{0.3}} &
                \col{\mstd{80.3}{0.8}} & \col{\mstd{70.2}{0.3}} & \col{\mstd{76.8}{0.7}} &
                \col{\mstd{78.5}{0.5}} & \col{\mstd{69.2}{0.3}} & \col{\mstd{76.9}{0.9}}  \\
            & ENGAGE & 
                \mstd{81.8}{0.4} & \mstd{72.4}{0.4} & \mstd{78.6}{0.5}& 
                \mstd{81.6}{0.2} & \mstd{72.2}{0.4} & \mstd{77.6}{0.5} & 
                \mstd{81.0}{1.1} & \mstd{71.7}{1.4} & \mstd{77.2}{1.7}\\ 
        \midrule
        \multicolumn{2}{l|}{\col{\xainode}} &
                \col{\mstd{\textbf{82.8}}{0.5}} & \col{\mstd{\textbf{74.0}}{0.4}} & \col{\mstd{\textbf{81.5}}{0.7}} &
                \col{\mstd{\textbf{82.6}}{0.5}} & \col{\mstd{\textbf{72.6}}{0.4}} & \col{\mstd{\textbf{80.7}}{0.5}} &
                \col{\mstd{\textbf{83.0}}{0.4}} & \col{\mstd{\textbf{73.0}}{0.6}} & \col{\mstd{\textbf{79.6}}{0.7}} \\          
        \bottomrule
    \end{tabular}
    }
    \caption{Node classification test set accuracy (in percentage). Mean and standard deviation over 5 runs with different initialization seeds. The best performing method is boldfaced.}
    \label{tab:acc}
\end{table}

\paragraph{{\bf R3: \xaidrop improves explainability}} Table~\ref{tab:expl} reports accuracy sufficiency (Eq.~\ref{eq:suf_norm_acc}) over the entire set of data (training, validation and test), again with mean and standard deviation over the 5 runs. As expected, XAI-based approaches improve explainability with respect to the baseline. It is interesting to highlight that dropping strategies are also quite effective in improving explainability, confirming the beneficial effect of dropping on training robustness. Notably, \xainode again stands out as the best performing method in all settings.
The improvement over the other XAI-based approaches highlights the effectiveness of using XAI as a dropping strategy in isolating the most relevant part of the input graphs. \rev{For additional xAI metrics analysis refer to Appendix~\ref{app:candidate_noisy_appendix}}.

\begin{table}[h!]
    \centering
    \resizebox{\textwidth}{!}{ 
        \begin{tabular}{l@{\hskip 9pt}l@{\hskip 5pt}|c@{\hskip 6pt}c@{\hskip 6pt}c@{\hskip 6pt}|c@{\hskip 6pt}c@{\hskip 6pt}c@{\hskip 6pt}|c@{\hskip 6pt}c@{\hskip 6pt}c}
        &        & \multicolumn{3}{c|}{GCN} & \multicolumn{3}{c|}{GAT} & \multicolumn{3}{c}{GIN}\\
        &Model   & Cora & CiteSeer & PubMed & Cora & CiteSeer & PubMed & Cora & CiteSeer & PubMed \\
        \toprule
            &Baseline & 
                \mstd{92.8}{0.3} &
                \mstd{88.3}{1.2} & 
                \mstd{92.9}{0.8}& 
                \mstd{90.6}{1.0} & 
                \mstd{89.6}{0.8} & 
                \mstd{92.5}{0.6}& 
                \mstd{91.3}{0.9} & 
                \mstd{88.8}{0.7} & 
                \mstd{92.4}{0.8}\\
        \midrule
        \multirow{4}{*}{{\rotatebox[origin=c]{90}{Random}}}
           & DropEdge & 
                \mstd{93.2}{0.7} & 
                \mstd{89.4}{1.7} & 
                \mstd{94.1}{1.2} &
                \mstd{90.2}{1.1} & 
                \mstd{90.9}{0.7} & 
                \mstd{94.2}{0.8}  &
                \mstd{90.2}{1.1} & 
                \mstd{90.9}{0.7} & 
                \mstd{93.8}{1.6}\\
            &\col{DropMess} & 
                \col{\mstd{92.9}{0.7}} &
                \col{\mstd{89.2}{0.5}} &
                \col{\mstd{92.7}{0.9}} &
                \col{\mstd{90.1}{0.4}} &
                \col{\mstd{91.5}{0.6}} & 
                \col{\mstd{92.9}{0.5}} &
                \col{\mstd{91.4}{0.5}} &
                \col{\mstd{91.1}{0.9}} & 
                \col{\mstd{92.5}{1.1}}\\
            & DropNode & 
                 \mstd{93.7}{0.4} & 
                 \mstd{90.9}{0.5} & 
                 \mstd{93.2}{0.7}& 
                 \mstd{92.9}{1.1} & 
                 \mstd{92.7}{0.8} & 
                 \mstd{93.5}{0.9} &
                \mstd{93.1}{1.6} &
                \mstd{92.7}{0.7} & 
                \mstd{93.0}{1.2}\\
            &\col{DropAggr} & 
                \col{\mstd{93.8}{1.1}} & 
                \col{\mstd{88.9}{1.2}} & 
                \col{\mstd{93.0}{0.8}} &
                \col{\mstd{90.6}{0.9}} & 
                \col{\mstd{89.9}{1.3}} &
                \col{\mstd{92.4}{0.9}} &
                \col{\mstd{91.2}{1.8}} &
                \col{\mstd{89.4}{1.4}} & 
                \col{\mstd{92.7}{0.9}}\\   
        \midrule
        \multirow{2}{*}{{\rotatebox[origin=c]{90}{Learn}}}
               & BBGDG &
               \mstd{89.2}{0.3}&
               \mstd{82.3}{0.4}&
               \mstd{84.2}{0.4}&
                -&
                -&
                -&
                -&-&-\\
         &    Learn2Drop & 
                \col{\mstd{90.1}{1.5}} & \col{\mstd{88.6}{1.8}} & 
                \col{\mstd{93.1}{1.0}} &
                \col{\mstd{88.4}{1.9}} & 
                \col{\mstd{87.7}{1.3}} & 
                \col{\mstd{92.8}{0.9}} &
                \col{\mstd{91.5}{1.0}} & 
                \col{\mstd{90.2}{1.2}} & 
                \col{\mstd{93.0}{0.9}} \\
        \midrule
        \multirow{3}{*}{{\rotatebox[origin=c]{90}{xAI}}} & 
            MATE & 
                \mstd{94.6}{0.7} &
                \mstd{92.1}{0.8} & 
                \mstd{92.9}{1.2} &

                \mstd{94.0}{1.4} & 
                \mstd{92.5}{0.9} & 
                \mstd{93.6}{1.0} &
                \mstd{94.0}{1.4} & 
                \mstd{92.5}{0.9} & 
                \mstd{93.2}{1.1} \\
   
            & \col{ExPass} & 
                \col{\mstd{92.8}{0.5}} & 
                \col{\mstd{90.6}{0.6}} & 
                \col{\mstd{93.9}{0.4}} &
                \col{\mstd{90.7}{0.3}} & 
                \col{\mstd{89.3}{0.8}} & 
                \col{\mstd{92.3}{0.4}} &
                \col{\mstd{90.9}{0.9}} & 
                \col{\mstd{89.9}{0.8}} & 
                \col{\mstd{92.7}{0.6}} \\
            & ENGAGE & 
                \mstd{92.9}{0.7} & 
                \mstd{90.7}{1.0} & 
                \mstd{94.3}{0.8}& 
                \mstd{90.2}{1.3} &
                \mstd{91.4}{1.1} & 
                \mstd{94.0}{0.5}& 
                
                \mstd{92.9}{1.4} &
                \mstd{92.0}{1.5} & 
                \mstd{94.2}{0.7}\\ 

        \midrule
        
        \multicolumn{2}{l|}{\col{\xainode}} &
                \col{\mstd{\textbf{97.2}}{0.6}} & 
                \col{\mstd{\textbf{95.2}}{0.7}} & 
                \col{\mstd{\textbf{97.3}}{0.9}} &
                \col{\mstd{\textbf{95.9}}{0.8}} & 
                \col{\mstd{\textbf{94.8}}{1.0}} & 
                \col{\mstd{\textbf{96.7}}{0.9}} &
                \col{\mstd{\textbf{96.4}}{1.0}} & 
                \col{\mstd{\textbf{95.5}}{1.3}} & 
                \col{\mstd{\textbf{97.0}}{0.5}}
                \\               
        \bottomrule
    \end{tabular}
    }
    \caption{Explainability of the different methods for node classification as measured by accuracy sufficiency.
    The best performing method is boldfaced.}
    \label{tab:expl}
\end{table}

\paragraph{{\bf R4: \xaidrop outperforms alternative strategies and improves explainability on link prediction tasks}} 
Tables~\ref{tab:accLP} and~\ref{tab:expl_link} report test set area under curve (AUC, a standard performance metric used in link prediction) and explainability (as measured by accuracy sufficiency) respectively, when \xaidrop is applied to link prediction tasks. Results confirm the generality of our XAI-driven dropping strategy, as \xaiedge outperforms all competitors (both dropping or XAI-based strategies) on all datasets, in terms of both prediction quality and explainability.

\begin{table}[h!]
    \centering
    \resizebox{\textwidth}{!}{ 
        \begin{tabular}{l@{\hskip 9pt}l@{\hskip 5pt}|c@{\hskip 6pt}c@{\hskip 6pt}c@{\hskip 6pt}|c@{\hskip 6pt}c@{\hskip 6pt}c@{\hskip 6pt}|c@{\hskip 6pt}c@{\hskip 6pt}c}
        &        & \multicolumn{3}{c|}{\textbf{GCN}} & \multicolumn{3}{c|}{\textbf{GAT}} & \multicolumn{3}{c}{\textbf{GIN}}\\
        &Model   & Cora & CiteS & PubM & Cora & CiteS & PubM & Cora & CiteS & PubM \\
        \toprule
            &Baseline & 
                \mstd{88.0}{1.0} & \mstd{86.7}{1.3} & \mstd{94.5}{0.2}& 
                \mstd{88.3}{1.1} & \mstd{85.6}{1.9} & \mstd{89.4}{0.3}& 
                \mstd{89.1}{1.3} & \mstd{87.0}{1.9} & \mstd{90.1}{0.5}\\
        \midrule
        \multirow{5}{*}{\rotatebox[origin=c]{90}{Random}} 
            & DropEdge & 
                \mstd{94.1}{0.7} & \mstd{90.5}{1.3} & \mstd{94.6}{0.3}& 
                \mstd{92.3}{0.3} & \mstd{94.6}{0.7} & \mstd{93.8}{0.7}& 
                \mstd{92.2}{1.0} & \mstd{91.9}{1.1} & \mstd{93.0}{0.9}\\
            &\col{DropMess} & 
                \col{\mstd{92.4}{0.9}} & \col{\mstd{90.8}{0.5}}& \col{\mstd{92.1}{0.8}} &
                \col{\mstd{92.1}{0.8}} & \col{\mstd{90.4}{0.7}}& \col{\mstd{91.5}{0.7}} &
                \col{\mstd{91.7}{0.9}} & \col{\mstd{91.2}{0.8}}& \col{\mstd{91.5}{1.6}} \\
            & DropNode & 
                \mstd{95.0}{0.8} & \mstd{91.4}{0.4} & \mstd{94.2}{0.8}& 
                \mstd{93.2}{1.3} & \mstd{90.7}{0.8} & \mstd{91.4}{0.5} & 
                \mstd{94.1}{1.2} & \mstd{92.8}{1.5} & \mstd{93.9}{1.3}\\
            &\col{DropAggr} & 
                \col{\mstd{90.5}{0.6}} & \col{\mstd{90.9}{0.5}} & \col{\mstd{92.3}{0.5}} &
                \col{\mstd{90.8}{0.4}} & \col{\mstd{90.3}{0.9}} & \col{\mstd{91.5}{0.8}} &
                \col{\mstd{90.5}{0.9}} & \col{\mstd{91.2}{1.1}} & \col{\mstd{91.4}{1.0}}\\ 
        \midrule
        \multirow{1}{*}{{\rotatebox[origin=c]{90}{L.}}} & 
            \col{Learn2Drop}  & 
                \mstd{89.6}{0.6} & \mstd{89.5}{0.9} & \mstd{90.3}{0.5}& 
                \mstd{90.1}{0.7} & \mstd{91.0}{1.2} & \mstd{92.1}{0.9}& 
                \mstd{90.2}{0.6} & \mstd{92.1}{1.6} & \mstd{91.6}{1.1}
            
              \\
        \midrule
        \multirow{4}{*}{\rotatebox[origin=c]{90}{xAI}} & 
            \col{FairDrop} &
                \col{\mstd{90.1}{0.7}} & \col{\mstd{88.4}{1.4}} & \col{\mstd{94.8}{0.2}} &
                \col{\mstd{87.8}{1.0}} & \col{\mstd{87.1}{1.1}} & \col{\mstd{87.1}{0.6}}&
                \col{\mstd{90.1}{1.2}} & \col{\mstd{89.3}{0.7}} & \col{\mstd{89.9}{0.9}} \\  
            & MATE & 
                \mstd{91.8}{0.6} & \mstd{90.4}{0.6} & \mstd{93.3}{0.9}& 
                \mstd{90.9}{0.8} & \mstd{88.2}{0.6} & \mstd{92.2}{0.7}& 
                \mstd{90.5}{1.2} & \mstd{86.9}{1.0} & \mstd{91.9}{1.2}\\     
            & \col{ExPass} & 
                \mstd{88.1}{1.3} & \mstd{87.2}{0.4} & \mstd{92.8}{0.9} &
                \mstd{88.5}{1.0} & \mstd{86.2}{0.7} & \mstd{92.0}{0.8} &
                \mstd{87.9}{1.3} & \mstd{86.4}{0.8} & \mstd{90.6}{0.5}\\
        \midrule
        \multicolumn{2}{l|}{\col{\xaiedge}} &
                \col{\mstd{\textbf{97.5}}{0.7}} & \col{\mstd{\textbf{98.6}}{0.9}} & \col{\mstd{\textbf{96.8}}{1.1}} &
                \col{\mstd{\textbf{96.8}}{1.0}} & \col{\mstd{\textbf{98.4}}{0.8}} & \col{\mstd{\textbf{95.9}}{0.9}}&
                \col{\mstd{\textbf{95.2}}{1.2}} & \col{\mstd{\textbf{94.8}}{0.5}} & \col{\mstd{\textbf{95.5}}{1.3}} \\          
        \bottomrule
    \end{tabular}
    }
    \caption{Test set AUC on link prediction, reported as the mean and standard deviation over 5 runs with different initialization seeds. The best performing method is boldfaced.}
    \label{tab:accLP}
\end{table}

\begin{table}[h!]
    \centering
    \resizebox{\textwidth}{!}{ 
        \begin{tabular}{l@{\hskip 9pt}l@{\hskip 5pt}|c@{\hskip 6pt}c@{\hskip 6pt}c@{\hskip 6pt}|c@{\hskip 6pt}c@{\hskip 6pt}c@{\hskip 6pt}|c@{\hskip 6pt}c@{\hskip 6pt}c}
        &        & \multicolumn{3}{c|}{GCN} & \multicolumn{3}{c|}{GAT} & \multicolumn{3}{c}{GIN}\\
        &Model   & Cora & CiteSeer & PubMed & Cora & CiteSeer & PubMed & Cora & CiteSeer & PubMed \\
        \toprule
            &Baseline & 
                \mstd{93.5}{0.8} &
                \mstd{91.6}{1.0} & 
                \mstd{94.0}{1.2}& 
                \mstd{92.9}{1.3} & 
                \mstd{90.8}{1.1} & 
                \mstd{93.8}{0.9}& 
                \mstd{93.0}{1.3} & 
                \mstd{92.1}{0.9} & 
                \mstd{93.0}{0.5}\\
        \midrule
        \multirow{4}{*}{{\rotatebox[origin=c]{90}{Random}}}
           & DropEdge & 
                \mstd{92.1}{1.1} & 
                \mstd{90.9}{1.5} & 
                \mstd{94.7}{1.2} &
                \mstd{92.3}{1.1} & 
                \mstd{90.9}{1.2} & 
                \mstd{94.5}{1.2}  &
                \mstd{92.4}{1.2} & 
                \mstd{90.9}{0.7} & 
                \mstd{93.9}{1.5}\\
            &\col{DropMess} & 
                \col{\mstd{93.1}{1.1}} &
                \col{\mstd{91.9}{0.9}} &
                \col{\mstd{93.6}{0.8}} &
                \col{\mstd{91.0}{0.5}} &
                \col{\mstd{91.5}{0.6}} & 
                \col{\mstd{93.9}{0.5}} &
                \col{\mstd{92.7}{1.0}} &
                \col{\mstd{92.3}{0.9}} & 
                \col{\mstd{92.9}{1.0}}\\
            & DropNode & 
                 \mstd{93.0}{0.9} & 
                 \mstd{91.9}{1.1} & 
                 \mstd{94.9}{1.2}& 
                 \mstd{93.1}{1.4} & 
                 \mstd{91.7}{1.7} & 
                 \mstd{92.8}{1.0} &
                \col{\mstd{92.8}{1.3}} &
                \col{\mstd{93.1}{1.0}} & 
                \col{\mstd{92.9}{1.2}}\\
            &\col{DropAggr} & 
                \col{\mstd{92.8}{1.1}} & 
                \col{\mstd{89.9}{1.3}} & 
                \col{\mstd{95.0}{0.6}} &
                \col{\mstd{91.3}{1.1}} & 
                \col{\mstd{90.9}{1.2}} &
                \col{\mstd{94.6}{1.1}} &
                \col{\mstd{92.2}{0.6}} &
                \col{\mstd{91.9}{1.6}} & 
                \col{\mstd{92.1}{0.7}}\\   
        \midrule
        \multirow{1}{*}{{\rotatebox[origin=c]{90}{L.}}}&   \col{Learn2Drop} & 
                \col{\mstd{91.9}{1.6}} & \col{\mstd{90.6}{1.2}} & 
                \col{\mstd{93.2}{1.0}} &
                \col{\mstd{92.4}{1.3}} & 
                \col{\mstd{87.7}{1.3}} & 
                \col{\mstd{92.1}{1.7}} &
                \col{\mstd{92.1}{1.3}} & 
                \col{\mstd{91.0}{1.4}} & 
                \col{\mstd{91.0}{0.6}} \\
        \midrule
        \multirow{3}{*}{{\rotatebox[origin=c]{90}{xAI}}} & 
             FairDrop & 
                \mstd{93.3}{0.9} & 
                \mstd{91.3}{1.0} & 
                \mstd{94.0}{0.9}& 
                \mstd{92.2}{0.5} &
                \mstd{91.9}{0.5} & 
                \mstd{91.4}{1.1}& 
                
                \mstd{92.5}{0.7} &
                \mstd{92.2}{1.2} & 
                \mstd{92.5}{0.9}\\ 
            & MATE & 
                \mstd{94.0}{0.9} &
                \mstd{92.5}{1.1} & 
                \mstd{94.3}{1.2} &

                \mstd{94.4}{1.2} & 
                \mstd{94.0}{0.8} & 
                \mstd{93.2}{1.1} &
                \mstd{94.2}{1.5} & 
                \mstd{93.5}{0.9} & 
                \mstd{94.2}{1.1} \\
   
            & \col{ExPass} & 
                \col{\mstd{94.2}{0.9}} & 
                \col{\mstd{92.2}{0.8}} & 
                \col{\mstd{92.6}{1.2}} &
                \col{\mstd{94.0}{1.1}} & 
                \col{\mstd{92.9}{1.0}} & 
                \col{\mstd{93.9}{0.4}} &
                \col{\mstd{91.2}{1.2}} & 
                \col{\mstd{90.5}{1.1}} & 
                \col{\mstd{92.1}{1.4}} \\

        \midrule
        
        \multicolumn{2}{l|}{\col{\xaiedge}} &
                \col{\mstd{\textbf{96.4}}{1.0}} & 
                \col{\mstd{\textbf{93.9}}{1.2}} & 
                \col{\mstd{\textbf{95.8}}{0.8}} &
                \col{\mstd{\textbf{95.3}}{1.2}} & 
                \col{\mstd{\textbf{94.4}}{1.3}} & 
                \col{\mstd{\textbf{95.2}}{0.9}} &
                \col{\mstd{\textbf{95.1}}{1.3}} & 
                \col{\mstd{\textbf{93.8}}{1.2}} & 
                \col{\mstd{\textbf{94.8}}{1.1}}
                \\               
        \bottomrule
    \end{tabular}
    }
    \caption{Explainability of the different methods for link prediction as measured by accuracy sufficiency. The best performing method is boldfaced.}
    \label{tab:expl_link}
\end{table}


\section{Conclusion}
\label{sec:concl}

In this work we introduced a simple XAI-based regularization framework for GNN training that selects nodes (for node classification) or edges (for link prediction) with highly confident predictions but poor explanations as candidates for dropping. Our experimental evaluation clearly showed that the proposed framework outperforms alternative dropping strategies as well as other XAI-based regularization techniques in terms of both accuracy and explainability. These promising results highlight the role of explainability-based regularization in improving training dynamics.
Future work include the exploration of the connection between explainability-based regularization and out-of-domain generalization, the application of similar XAI-based solutions to design augmentation strategies, and the study of explainability-based dropping for other classes of deep learning architectures.

\bibliographystyle{unsrtnat}
\bibliography{reference}

\appendix




\section{Dataset Statistics}\label{app:data}
Table \ref{tab:graph_stat} outlines the key characteristics of the dataset, including the number of classes, nodes, \rev{and unidirectional edges in the networks}.
\begin{table}[h]
    \centering
    \begin{tabular}{l|c|r|r}
                    & \# classes & \# nodes & \# edges \\
        \toprule
         CiteSeer   & 6 & 3,327 &  9,104 \\
         Cora       & 7 & 2,708 &  10,556\\
         PubMed     & 3 & 19,717 & 88,648\\
         \rev{OGBN-Arxiv} &\rev{40} & \rev{169,343} & \rev{2,332,486}\\
    \end{tabular}
    \caption{Dataset statistics.}
    \label{tab:graph_stat}
\end{table}

\section{Additional results}
\label{app:literature_res}

The computation of explainability for different architectures, methods, and datasets requires access to the weights of the model. To achieve this goal, we have retrained the models with a specific regularization method on each, in case the authors do not release the weights of the models. For these motivations, we have reported in Table~\ref{tab:acc} the accuracy that we get by retraining from scratch the model of interest with the hyperparameters written in the paper or source and in case of missing hyperparameters information, by applying grid search during hyperparameter optimization.
Table \ref{tab:original_accuracy} reports for each method the results from the corresponding original paper as reported under the source column, when available. Results confirm the advantage of \xainode over all alternatives.

\begin{table}[h!]
    \centering
    
    \resizebox{\textwidth}{!}{ 
    \begin{tabular}{ll|c|ccc|ccc}&        
       & & \multicolumn{3}{c|}{GCN} & \multicolumn{3}{c}{GAT}   \\
        &Model   
        &Source
        & Cora
        & CiteSeer 
        & PubMed 
        & Cora
        & CiteSeer
        & PubMed \\
        \toprule
            &Baseline & 
                \cite{fang2023dropmessage}&
                \mstd{80.7}{0.4} & \mstd{70.8}{0.5} & \mstd{75.9}{0.7}& 
                \mstd{81.4}{0.6} & \mstd{70.1}{0.6} & \mstd{77.2}{0.5}\\
        \midrule
        \multirow{4}{*}{Random}
            & DropEdge& 
                \cite{fang2023dropmessage}&
               \mstd{81.7}{0.9} & \mstd{71.4}{0.7} & \mstd{79.1}{0.8}& 
                \mstd{81.8}{0.8} & \mstd{71.1}{1.0} & \mstd{77.7}{0.8}\\
            &\col{DropMess} & 
                \col{\cite{fang2023dropmessage}}&
                \col{\mstd{83.3}{0.6}} & \col{\mstd{71.8}{0.6}} & \col{\mstd{79.2}{0.5}} &
                \col{\mstd{82.2}{0.7}} & \col{\mstd{71.5}{0.7}} & \col{\mstd{78.1}{0.5}} \\
            & DropNode & 
                \cite{feng2020graph}&
                \mstd{84.5}{0.4} & \mstd{74.2}{0.3} & \mstd{80.0}{0.3}& 
                \mstd{84.3}{0.4} & \mstd{73.2}{0.4} & \mstd{79.2}{0.6}\\
            &\col{DropAggr} & 
                \col{\cite{jiang2023dropagg}}&
                \col{83.1} & \col{72.8} & \col{-} &
                \col{\mstd{83.6}{0.7}} & \col{\mstd{72.9}{0.5}} & \col{-} \\   
        \midrule
        \multirow{2}{*}{Learning} & 
            \col{Learn2Drop} & 
                \col{\cite{luo2021learning}}&
                \col{\mstd{82.8}{0.3}} & \col{\mstd{72.7}{0.2}} & \col{\mstd{79.8}{0.2}} &
                \col{\mstd{84.4}{0.2}} & \col{\mstd{73.7}{0.3}} & \col{\mstd{79.3}{0.1}} \\
            &BBGDC &\cite{hasanzadeh2020bayesian}
            &\mstd{81.8}{1.0} & \mstd{71.5}{0.6} & - & - & - & -\\
        \midrule   
        \multirow{2}{*}{xAI-Based} &  \col{ExPass} & 
        \col{\cite{giunchiglia2022towards}}
                 & 
                \col{-} & \col{-} & \col{\mstd{76.2}{0.3}} &
                \col{-} & \col{-} & \col{-} \\
            & ENGAGE & 
            \cite{shi2023ENGAGE}
                &
                \mstd{84.1}{0.2} & \mstd{72.4}{0.5} & - & 
                \mstd{83.8}{0.5} & \mstd{72.4}{0.5} & -\\ 
        \midrule
        \multicolumn{2}{l}{\xainode} & 
                &
                \mstd{\textbf{84.7}}{0.7} & \mstd{\textbf{74.6}}{0.9} & \mstd{\textbf{82.0}}{0.9}& 
                \mstd{\textbf{84.5}}{0.6} & \mstd{\textbf{73.6}}{0.6} & \mstd{\textbf{81.2}}{0.8}\\             
        \bottomrule
    \end{tabular}
    }
    \caption{Test set accuracy on GCN (in percentage). The number of runs for computing standard deviations, when available, can be found in the corresponding paper reported under the "Source" column.}
    \label{tab:original_accuracy}
\end{table}

\section{Approximated saliency map}\label{app:aproxSal}
Saliency map is an explainer that produces an importance score for each node feature given a single model prediction.
In general, saliency map is applied to a single node $v$ to get the local explanation in the k-hop neighborhood of the node of interest $v$.
For computational efficiency, rather than applying one forward step for each candidate node, we compute a single forward step for the entire set of candidate nodes.
The node feature importance will then be the gradient of a single forward step on the entire set of candidate nodes, rather than the gradient of a forward step on just one candidate node.
Once we have obtained the node feature importance, the aggregation of them has been considered as node importance score.
These node scores are then used to generate the explanation subgraph where the top-$K$\% most important neighboring nodes are retained.
This means that a batch of candidate nodes is associated with a single explanation subgraph rather than associating one different explanation subgraph with each single candidate node in the batch.

\rev{This approximated variant of the Saliency Map method dramatically reduces the time complexity of the approach.
Let's consider the worst case scenario in the transductive node classification task: all the nodes'  predictions have a confidence score  (Equation \ref{eq:conf}) higher than the threshold confidence $C(v) > \theta, \forall v \in \mathcal{V}$, so that all the nodes are candidate noisy nodes $v \in V', \forall v \in \mathcal{G}$. Then $n$ different explanations have to be computed, with $n=|\mathcal{V}|$ being the number of nodes in the graph. In this setting, the standard Saliency Map procedure would require $n$ different forward steps, one for each node: $f_{v}(G); \forall v \in \mathcal{V}$ to compute $n$ different local explanations. On the contrary, Approximated Saliency Map applies a single forward step on the entire input graph $f(G)$, and the prediction is backpropagated to leverage the gradient of each feature in the input graph as the explanation importance score. Finally, these feature-level importance scores are averaged at the node level to get the importance of each node. 
Once we get these importance scores, for each node, the connections with the $\theta \%$-most relevant neighboring nodes are retained, while the incident edges from the not-relevant nodes are dropped to prevent the propagation of direct incoming messages during the forward step. This simple approximation reduces also the computational costs required to create $n$ different explanation subgraphs. 
Once the explanation graph, which retains only the most important connections according to the Approximated Saliency Map explainer, has been isolated, a single forward step is used to compute the prediction for all the nodes in the graph having the original topology $f(G)$, and another forward step computes the prediction for all the nodes in the graph having the explanation topology $f(G_{exp})_{V'}$.
Finally, the Kullback-Lieber (KL) divergence sufficiency score $KL_{suf}(v)$(Equation \ref{eq:kl-div}) is computed, for each node in the set of candidate noisy nodes $v \in V'$, between the two probability distributions obtained by feeding the model with the original graph and the explanation subgraph respectively.} 

Wrapping up, the overhead introduced by \xaidrop in the case of node classification consists of:
\begin{itemize}
    \item performing (only) one forward step to compute the explanations, regardless of the number of nodes in the set of candidate noisy nodes;
    \item performing (only) two forward steps to compute the predictions for the original graph and the explanation subgraph;
    \item computing the KL-divergence.
\end{itemize}

\section{\rev{Explainer comparison}}\label{app:other_explainers}

\begin{figure}[h!]
    \centering
    \includegraphics[width=0.7\linewidth]{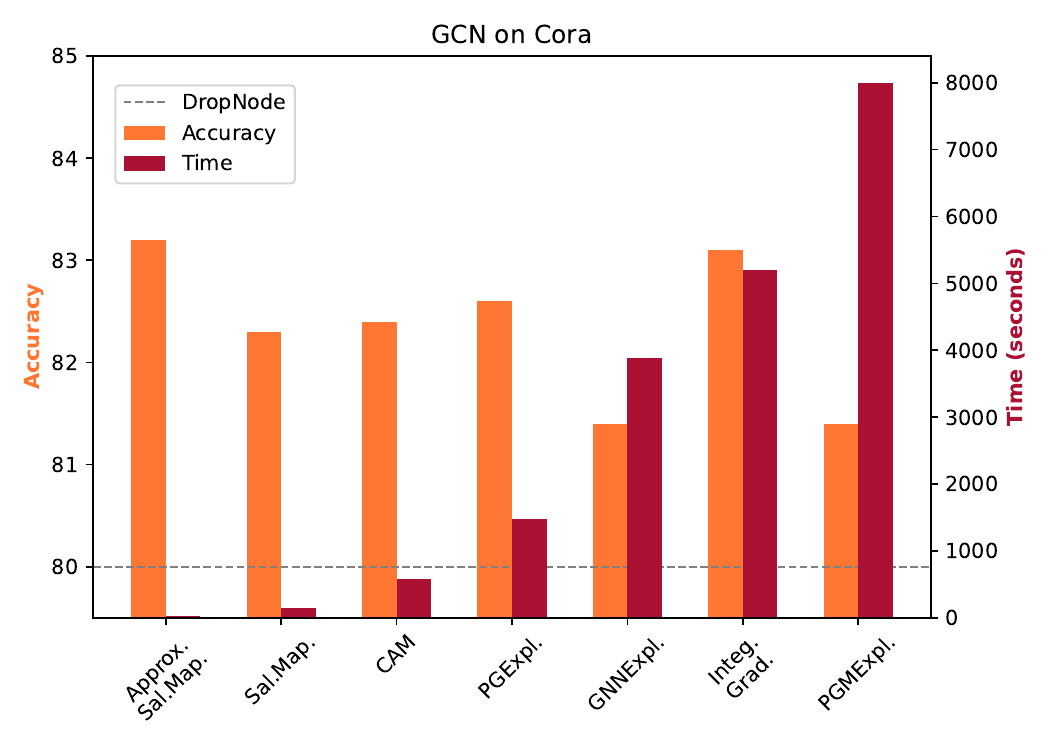}
    \caption{\rev{Test accuracy (left axis) and training time (right axis) when using different explainers for \xainode applied on Cora for node classification with GCN architecture. The dotted line represents the accuracy achieved when using the baseline DropNode random strategy.}}
    \label{fig:other_explainers_fig}
\end{figure}

\rev{In this section, we explore the flexibility of \xaidrop to the usage of alternative explainers with respect to the Approximated Saliency Map used in the main paper. In Figure \ref{fig:other_explainers_fig}, we present the performance and training time of our method using different explainers in terms of accuracy (left axis) and training time (right axis).
For this ablation study, we have explored different types of explainers:}
\begin{itemize}
    \item \rev{Gradient-based (Integrated Gradients~\cite{sundararajan2017axiomatic}, Saliency-Map~\cite{simonyan2013deep})}
    \item \rev{Perturbation-based (GNNExplainer~\cite{ying2019gnnexplainer}, PGExplainer~\cite{luo2020parameterized})}
    \item \rev{Decomposition-based (CAM~\cite{baldassarre2019explainability})}
    \item \rev{Surrogate-based (PGM-Explainer~\cite{vu2020pgm})}
\end{itemize}

\rev{Quite remarkably, \xaidrop manages to improve performance with respect to the random node-dropping strategy {\em regardless of the explainer being used}. On the other hand, this plot highlights the substantial computational advantage of Approximated Saliency Map (and to a less extent Saliency Map and CAM) over more complex alternatives, without incurring in a reduction of generalization capabilities.}

\section{Computational complexity}
\label{app:time}

\begin{figure}[h!]
    \centering
    \includegraphics[width=\textwidth]{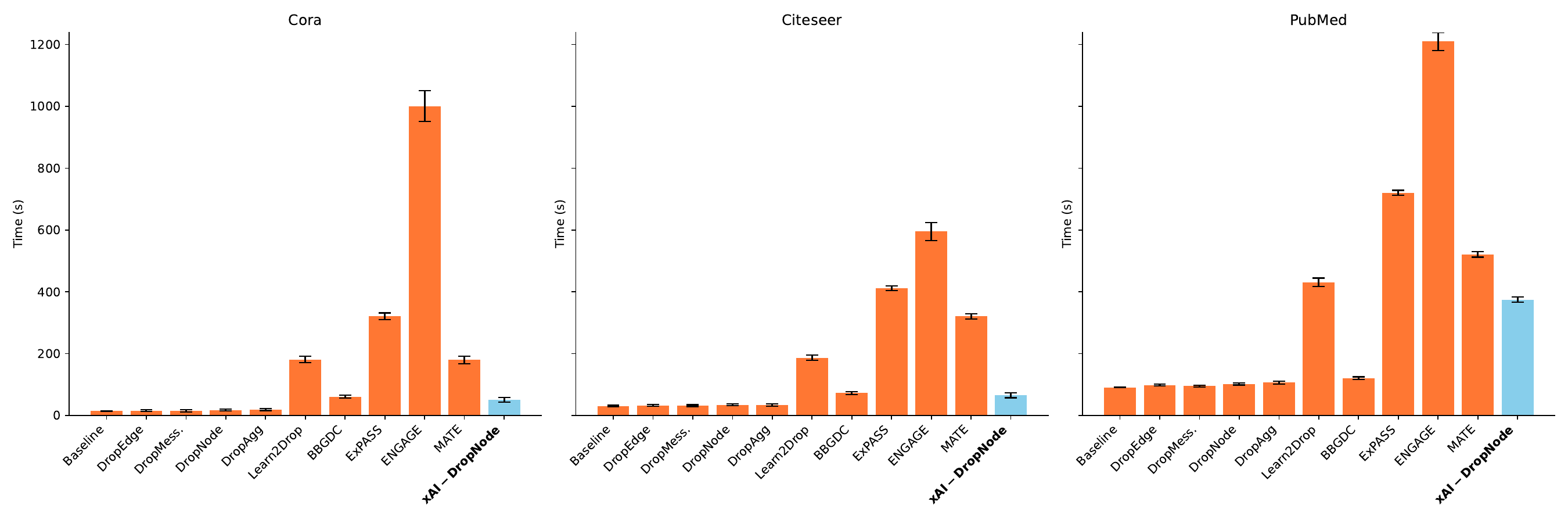}
    \caption{\revsecond{The histogram of the time in seconds required for training GCN on Cora, Citeseer, and PubMed with each regularization method used for node classification.}}
    \label{fig:training-time}
\end{figure}

Training time is a crucial challenge in designing topological regularizers for GNNs.
In Figure \ref{fig:training-time}, we report the training time required for each method by iterating training for the same number of epochs when using Cora as dataset and GCN as architecture.
While baseline and random methods (i.e. DropEdge, DropMessage, DropNode, DropAgg) are extremely fast, and almost have the same training time, the additional operations computed in other dropping strategies, introduce additional overhead.
In Figure \ref{fig:training-time} it is clear that there are strategies whose computational overhead is crucial in analyzing their performance.
ExPass introduces a relevant overhead due to the usage of GNNExplainer for producing explanations. GNNExplainer is a computationally demanding explainer because it requires a separate iterative learning procedure and applying it to the input graph dramatically slows down the training procedure.
Also MATE, Learn2Drop, and ENGAGE require a double training procedure.
MATE introduces a Meta-learning approach and is a model that requires more parameters than the traditional dropping procedure to stabilize its training dynamics.
Learn2Drop, apart from training the model for node classification, needs to train the model to learn a denoised topology and this objective requires a lot of parameters.
ENGAGE incorporates an unsupervised step for learning robust embeddings by optimizing a contrastive objective and a supervised step for doing prediction on top of these embeddings. 
The unsupervised step requires deep, highly non-linear functions and many parameters to learn to be effective.
Our approach \xainode, despite the introduced overhead, thanks to the approximated variant of Saliency Map, defined in Appendix~\ref{app:aproxSal}, and the Node selection based on confidence, still guarantees a manageable training time.

\begin{figure}[t!]
    \centering
    \includegraphics[width=\textwidth]{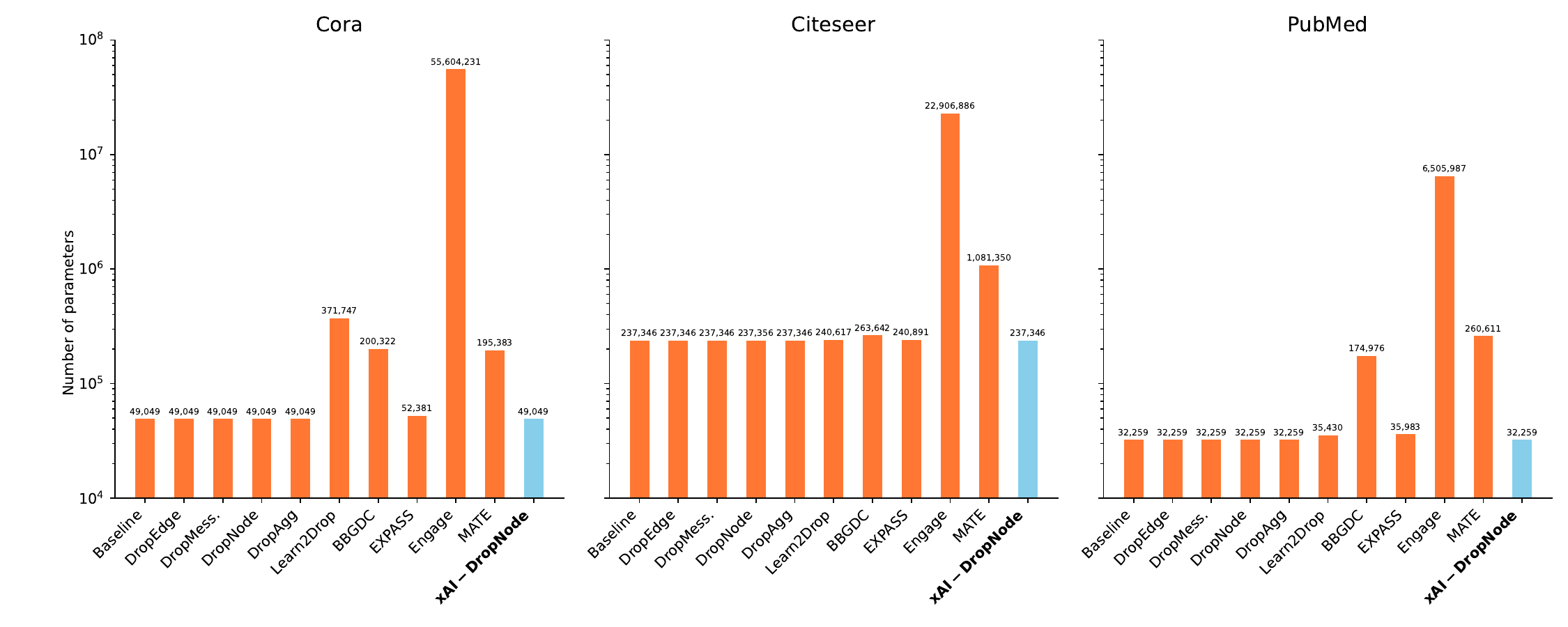}
    \caption{\revsecond{The log-scale histogram of the parameters used for training GCN for node classification task with Cora, Citeseer, and Pubmed datasets.}}
    \label{fig:all-parameters}
\end{figure}

In the analysis of the computational complexity, the number of parameters required by each method also plays a crucial role. As in Figure \ref{fig:training-time} the analysis is conducted on the Cora dataset trained with GCN.
In Figure \ref{fig:all-parameters}, the number of parameters is plotted with a log-scale histogram rather than a linear-scale histogram due to the huge amount of parameters required by ENGAGE for the unsupervised training procedure, before the finetuning stage. Other methods which requires much more parameters than the other strategies are Learn2Drop and MATE, because of the need for parameters to, respectively, optimize the robustness of the topology and the Meta-Learning inner stage.
On the contrary, the released version of BBGDC simply uses a wider hidden layer.
Finally, the baseline, random drop strategies (DropEdge, DropMess, DropNode, DropAgg), and xAI-guided methods (ExPass and xAI-DropNode) use the same hidden size and the same network depth. The small number of additional parameters introduced by ExPass is due to the choice of a parametric explainer.
Figure \ref{fig:all-parameters} confirms, as Figure \ref{fig:training-time}, that \xainode does not introduce a meaningful computational overhead with respect to other random dropping strategies; and at the meantime, \xainode exhibits an evident computational advantage with respect to learnable and alternative xAI-based dropping strategies.

\section{\rev{Hyperparameter sensitivity}}\label{hyperparameter_and_sensitivity}
\subsection{\rev{Confidence}}
\rev{One of the most important hyperparameters in our method is the confident threshold $\theta \in [0,1]$. This hyperparameter is necessary to decide whether a node is a candidate noisy node or not. To fully comprehend its rule, we can start by analyzing the two extreme cases:
\begin{itemize}
    \item if $\theta=0$: all nodes in the graph are candidate noisy nodes, regardless of their confidence.
    The consequence is that the dropping probability of each node will be biased exclusively based on its explanation quality.
    \item if $\theta=1$: no node has a confidence larger than the threshold, and the strategy boils down to random dropping.
    \end{itemize}}

\begin{figure}[h]
    \centering
    \includegraphics[width=\textwidth]{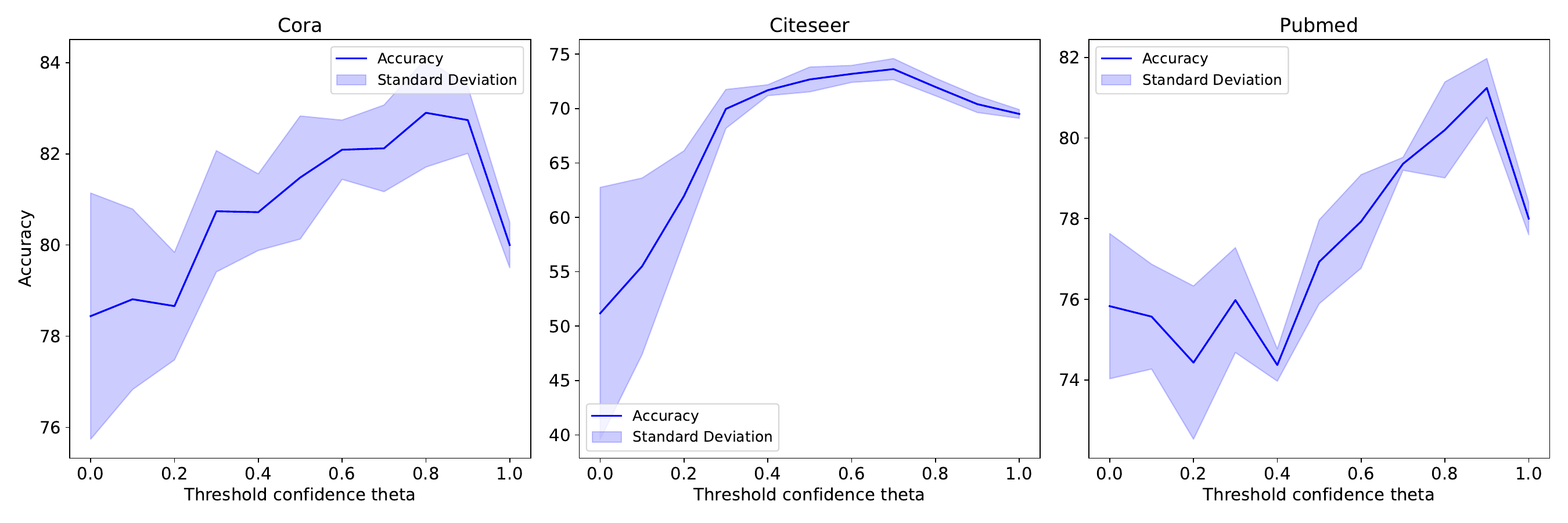}
    \caption{\rev{Test Accuracy and standard deviations on some node classification datasets (Cora, Citeseer, and Pubmed) trained with GCN varying confidence threshold $\theta$.}}
    \label{fig:threshold-tuning}
\end{figure}

\rev{In Figure~\ref{fig:threshold-tuning}, we have reported how the tuning of the confidence hyperparameter $\theta$ affects test accuracy on Cora, Citeseer, and Pubmed trained with GCN for Node Classification task. Results show that both completely explainability-guided ($\theta=0)$ and completely random ($\theta=1)$ strategies are suboptimal, and that a threshold around 0.8 is reasonable for all datasets.}

\subsection{\rev{Dropping probability}}
\begin{figure}[h]
    \centering
    \includegraphics[width=\textwidth]{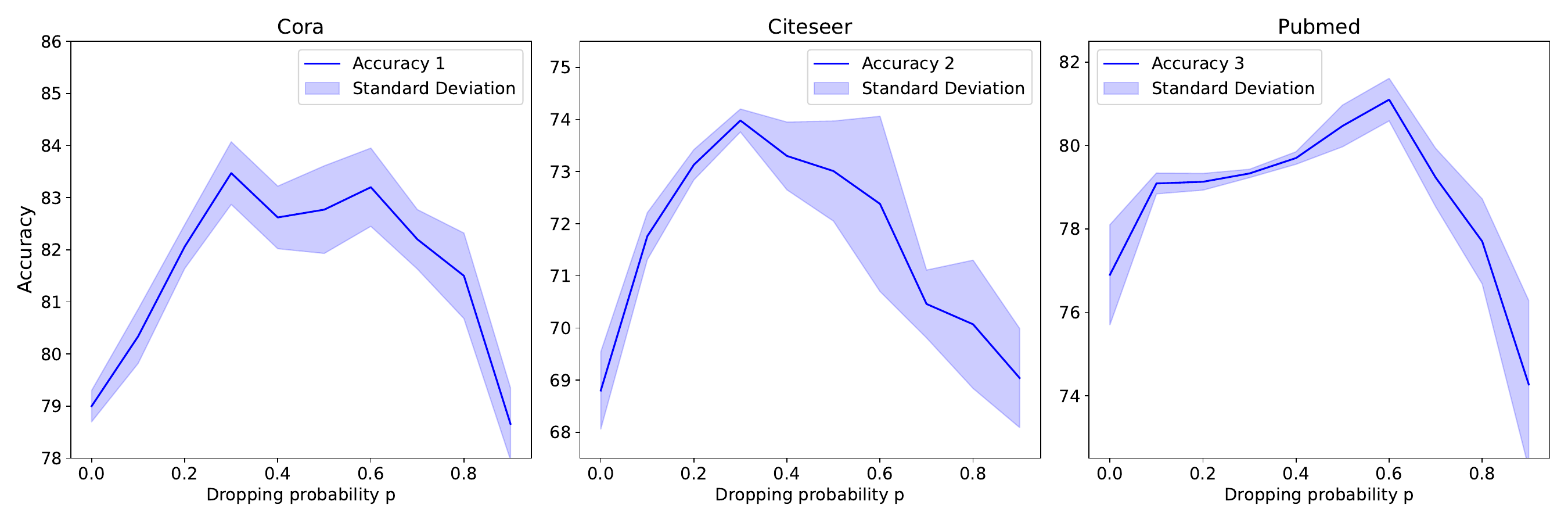}
    \caption{\rev{Test accuracy on Node classification with GCN on multiple datasets (Cora, Citeseer, Pubmed) varying dropping probability $\theta$.}}
    \label{fig:hyper-dropping-probability}
\end{figure}

\rev{The dropping probability $p$ is a crucial hyperparameter for properly applying \xaidrop. As with any dropping strategies, the tuning of this hyperparameter strongly depends on the input graph, and is also related to the design of the GNN. In Figure~\ref{fig:hyper-dropping-probability} we show how the test accuracy varies when changing the dropping probability. From this empirical evidence, we note that larger datasets such as Pubmed have better results for larger values of $p$, i.e., with a more aggressive dropping. Furthermore, it is interesting to notice that removing the $90\%$ of the input nodes leads to results similar to the baseline, which does not apply any dropping.}

\section{Evolution of node confidence and explainability, ablation studies}\label{app:candidate_noisy_appendix}

\begin{figure}[h!]
    \centering
    \includegraphics[width=\textwidth]{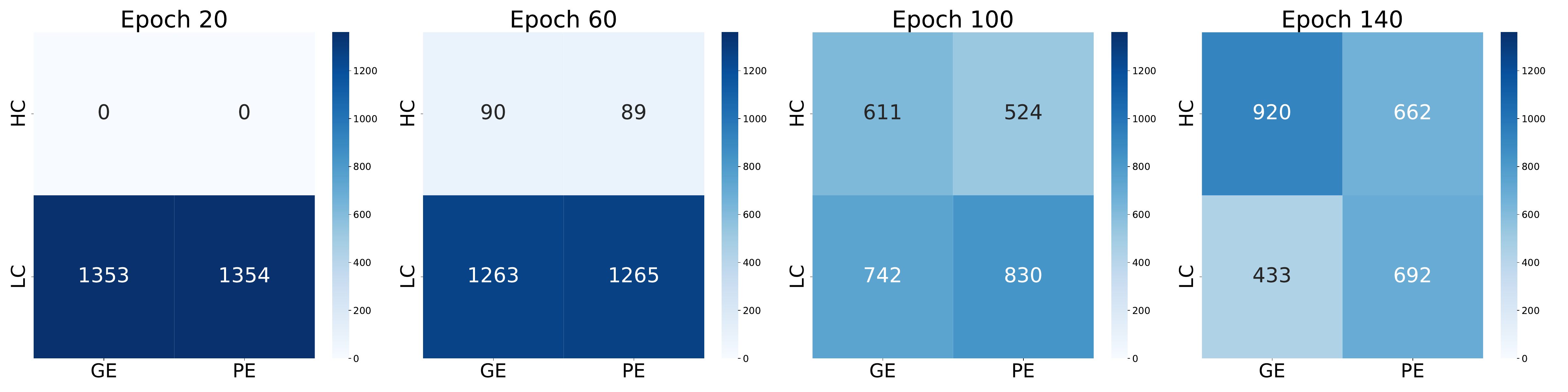}
    \caption{\rev{Confusion matrices for an increasing number of training epochs, showing nodes with high confidence and good explanations (HC-GE), high confidence and poor explanations (HC-PE),  low confidence and good explanations (LC-GE) and low confidence and poor explanations (LC-PE).}}
   \label{fig:xAI-overtime}
\end{figure}

\rev{The \xaidrop method relies on a crucial intuition: the combination of confidence and explanation quality can be used as a proxy for pinpointing harmful nodes in a graph, the removal of which stabilizes training. To better show how training evolves using xAI-Drop, Figure \ref{fig:xAI-overtime} reports a series of confusion matrices (at different stages of training) reporting nodes with high confidence and good explanations (HC-GE), high confidence and poor explanations (HC-PE),  low confidence and good explanations (LC-GE) and low confidence and poor explanations (LC-PE). Results show how most nodes have initially low confidence. Thanks to training, the confidence of nodes increases, but high-confidence nodes are equally distributed among good-explanation and poor-explanation ones. While training progresses, increasingly more nodes have high confidence and good explanations, as expected.}

\begin{figure}
    \centering
    \includegraphics[width=\textwidth]{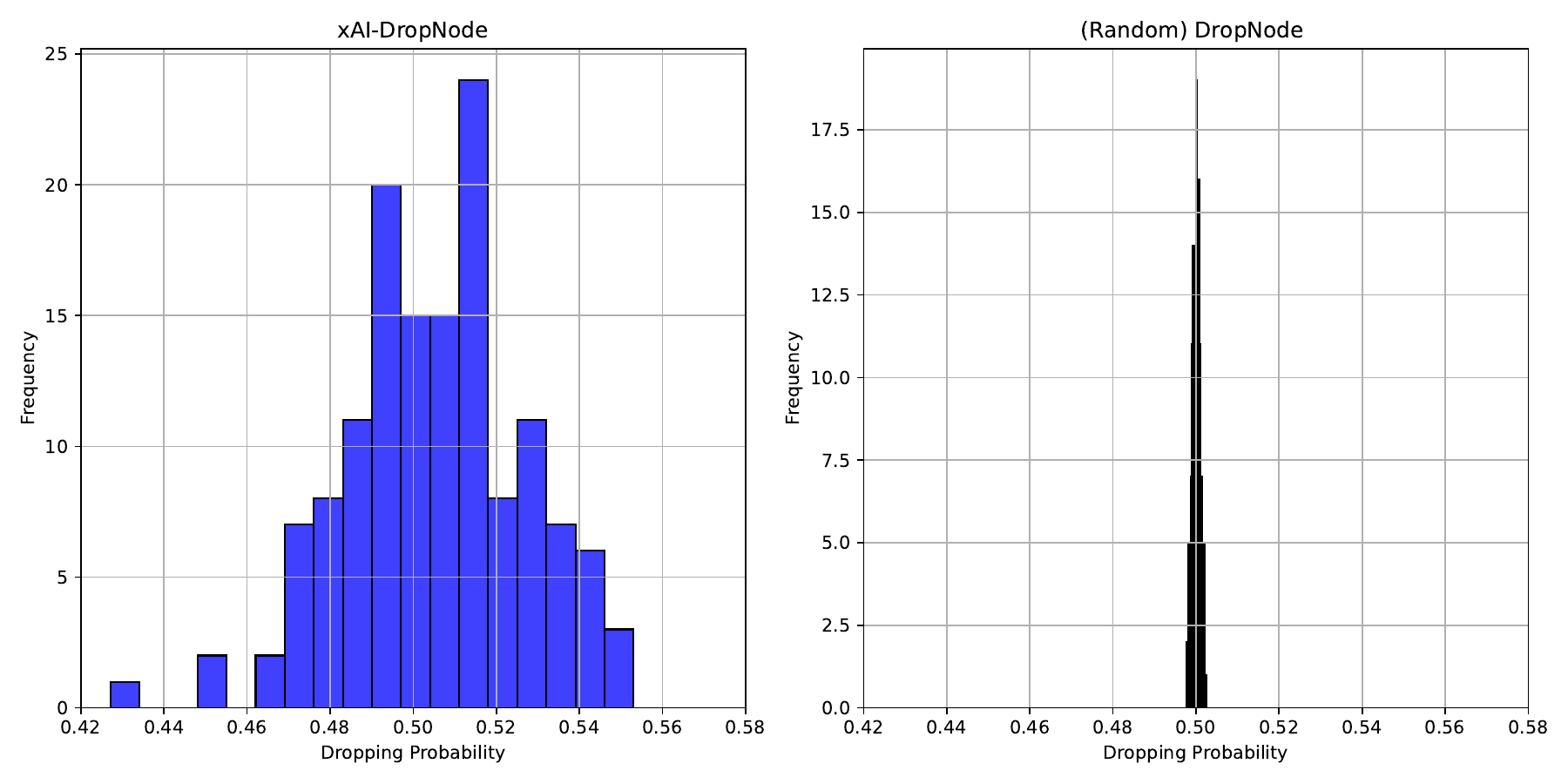}
    \caption{\rev{Histograms showing the average dropping probability of each node in a graph computed over all the training epochs for \xainode (left) and \dropnode (right) respectively.}}   \label{fig:drop-prob-over-time}
\end{figure}

\rev{Figure~\ref{fig:drop-prob-over-time} shows the histogram of the dropping probability of a node averaged over the set of training epochs. Clearly, the histogram converges to a Delta Dirac (on $p=0.5$) for the random strategy (\dropnode, right plot), which corresponds to a uniform dropping probability for all nodes. On the contrary, \xaidrop (left histogram) significantly biases the behaviour of nodes over training, so that part of the nodes is consistently identified as harmful (often dropped during training) or beneficial (mostly retained during training), stabilizing training.}

\rev{Finally, Table \ref{tab:noisy-strategies} presents the results of an ablation study where we altered the dropping strategy. Alternatives explored include confidence-only (high or low), explanation-only (good or poor) and their combinations. Results clearly indicate the advantage of the \xaidrop strategy focusing on high-confidence, poorly explained cases. It is important to highlight that dropping low confidence nodes is especially detrimental, most likely because it destabilizes training removing instances that still need to be properly learned.} 

\begin{table}[h!]
    \centering{ 
        \begin{tabular}{c|c}
             \textbf{Noisy node selection Criterion} & \textbf{Cora} \\
            \toprule
              \col{Random}& \col{\mstd{80.0}{0.5}} \\
              HighConfidence & \mstd{80.6}{0.4} \\
              \col{LowConfidence} & \col{\mstd{74.5}{1.5}}  \\
              LowConfidence+PoorXAI & \mstd{78.4}{0.9} \\
              \col{HighConfidence+GoodXAI} & \col{\mstd{79.8}{0.5}}  \\
              LowConfidence+GoodXAI & \mstd{77.4}{1.6} \\  
              \col{LowConfidence+Random} &
              \col{\mstd{76.9}{1.4}} \\
              HighConfidence+Random & \mstd{81.2}{0.9} \\     
              \col{PoorXAI} & \col{\mstd{80.4}{0.9}} \\
              GoodXAI & \mstd{79.5}{0.5} \\
              \col{\textbf{xAI-Drop}} & \col{\mstd{\textbf{82.8}}{\textbf{0.5}}} \\ 
            \bottomrule
        \end{tabular}
    }    
    \caption{\rev{Test set accuracy (in percentage) on Cora dataset for node classification trained with GCN by comparing different metrics for identifying noisy nodes. The standard deviation is computed over three runs.}}
    \label{tab:noisy-strategies}
\end{table}

\section{\rev{Post-Hoc Explanation evaluation across explanations metrics}}\label{app:xai-assessment-app}

\rev{In this section we evaluate the quality of the explanations obtained using \xaidrop and its competitors (with a GCN) in terms of alternative explanation quality metrics, namely KL-Necessity and KL-Sufficiency.} 

\rev{KL-Sufficiency follows the definition of sufficiency described in Equation \ref{eq:suf}, where the distance criterion is the Kullback-Lieber divergence as reported in \ref{eq:kl-div} between the two probability distributions($(f_{v}(G)_{i})$, $f_{v}(G_{exp}(v))$) produced by feeding the GNN, respectively, the original graph $G$ and the explanation subgraph $G_{exp}(v)$ for the node $v \in G$.}

\begin{equation}\label{eq:kl-div}
KL_{suf}(v) = \sum_{i=1}^{c}\mathcal(f_{v}(G))_{i} \; log\left(\frac{(f_{v}(G))_{i}} {f_{v}(G_{exp}(v))_{i}}\right)
\end{equation}

\rev{KL-Necessity, on the other hand, removes the explanation from the neighborhood of the node of interest, to define whether the explanation is necessary for producing the same prediction. It is computed as the KL-distance between the probability distributions produced by feeding the entire graph $f_{v}(G)_{i}$ and the non-relevant subgraph $f_{v}(G \setminus G_{exp}(v))$:}

\begin{equation}\label{eq:kl-nec}
KL_{nec}(v) = \sum_{i=1}^{c}\mathcal(f_{v}(G))_{i} \; log\left(\frac{(f_{v}(G))_{i}} {f_{v}(G \setminus G_{exp}(v))_{i}}\right)
\end{equation}

\begin{figure}[h]
    \centering
    \includegraphics[width=\textwidth]{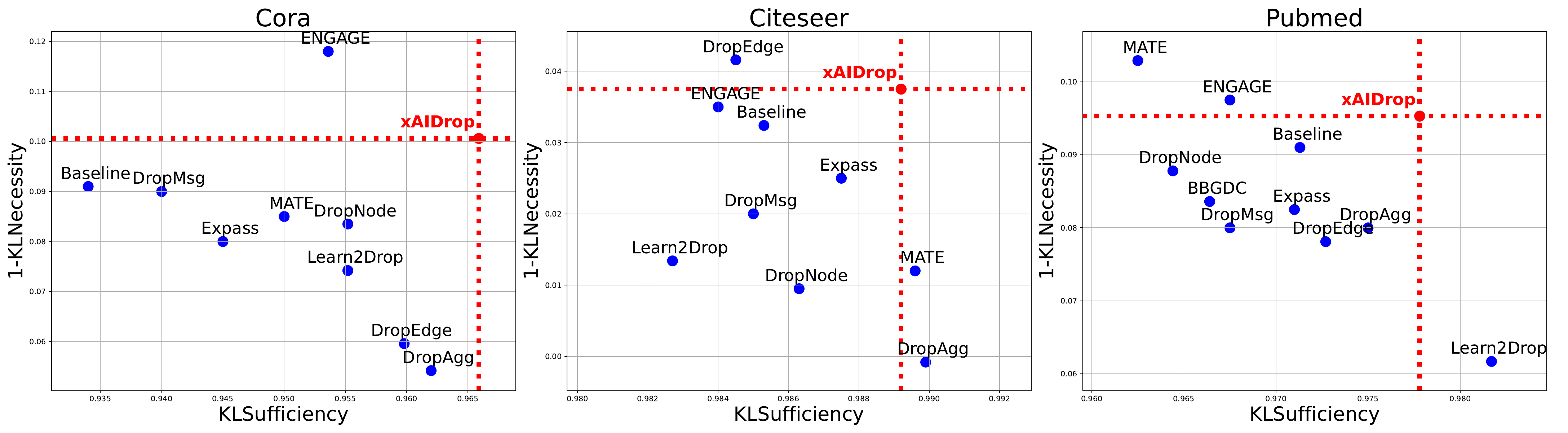}
    \caption{\rev{Scatter plot representing the quality of the explanations produced through Saliency Map on GCN across multiple datasets (Cora, Citeseer, Pubmed), measured in terms of KL-Sufficiency (x-axis) and 1- KL-Necessity (y-axis).}}
    \label{fig:additional-xai}
\end{figure}

\rev{Figure~\ref{fig:additional-xai} reports results of the different methods in terms of KL-Sufficiency {\em and} KL-Necessity, as only a reasonable trade-off between the two is an indicator of a good quality explanation. Results clearly indicate that \xaidrop scores the best trade-off between the two metrics, thus achieving the best explanations for all datasets.}

\section{\rev{xAI-DropEdge}}
\label{app:dropedge}

\rev{The overall algorithm for \xaiedge is outlined in Algorithm~\ref{algo:algo-xaiedge}. As in Algorithm~\ref{algo:algo1}, the algorithm takes as input a graph $\mathcal{G}$, the GNN architecture to be trained $f$, and the hyperparameters $\theta$ and $p$. In each epoch, the algorithm from the entire set of edges $\mathcal{E}$ selects the edges $\mathcal{E'}$ with a prediction confidence score higher than the confidence threshold $\theta$. Explanations qualities for all edges in $\mathcal{E'}$ are assessed via the Fidelity sufficiency score $F_{suf}$ (Equation 
\ref{eq:suf}). These explanation scores are mapped in probabilities through the Yao-Johnson mapping as described in Equation \ref{eq:yeo-johnson}, as happens for nodes in the xAI-DropNode variant.}
%
\rev{Once the biased dropping probabilities $p(v)$ have been computed, \xaiedge removes edges $e \in \mathcal{E}$ from the edge set $\mathcal{E}$ based on a edge dropping mask $B^{\mathcal{E}} \in \{0,1\} ^ {|\mathcal{V}| \times |\mathcal{V}|}$ defined as follows:}

\begin{equation}\label{eq:xai-drop_edge}
B^{\mathcal{E}}_{i,j} \sim Bernoulli(1-p((i,j)))
\end{equation}

\rev{where $p((i,j)) = 1$ if $(i,j) \notin \mathcal{E}$. The edge-dropping operation can be compactly represented in terms of Hadamard product between the binary edge dropping mask $B^{\mathcal{E}}$ and the adjacency matrix of the input graph $A$:}

\begin{equation}
\label{eq:dropedge_adj}
A' = A \otimes B^{\mathcal{E}}
\end{equation}


\begin{algorithm}[h!]
\caption{\rev{\xaidrop algorithm for link prediction. $G=(\mathcal{V},\mathcal{E},\textbf{X}_\mathcal{V},\textbf{X}_\mathcal{E})$ is a graph, $f$ is the GNN, $\theta, p$ are hyper-parameters}}
\label{algo:algo-xaiedge}
\begin{algorithmic}[1]
\Procedure{\xaidrop}{$G=(\mathcal{V},\mathcal{E},\textbf{X}_\mathcal{V},\textbf{X}_\mathcal{E})$,$f$, $\theta$, $p$}
    \For{$e \in \text{Epochs}$}
        \State $\mathcal{E}' \leftarrow \textsc{highest-confidence}(G,\mathcal{E},f, \theta$) \Comment{Equation \ref{eq:conf}} 
        \For{$e \in \mathcal{E}'$}
            \State $G_{exp}(e) \leftarrow \textsc{saliency-map}(G,e)$  \Comment{Equation \ref{eq:sal}}
             \State $F_{suf}(e) \gets \textsc{fidelity}(f,G,G_{exp}(e))$ \Comment{Equation \ref{eq:suf}}
        \EndFor
        \State $\mathbf{p} \leftarrow \textsc{dropping-probabilities}(F_{suf}, p)$
        \Comment{Equation \ref{eq:xai-drop_edge}}
        \State $A' \rightarrow \xaiedge(G,\mathbf{p})$ \Comment{Equation \ref{eq:dropedge_adj}}
        \State $f \gets \textsc{train}(f, G, A')$
    \EndFor
    \State $\mathcal{Y} \gets \textsc{evaluate}(f, G, A)$
\EndProcedure
\end{algorithmic}
\end{algorithm}

\newpage
\section{Scaling xAI-Drop}\label{app:scale-xai-drop}

\rev{Dropping strategies, apart from the advantages analysed in Section~\ref{sec:exps}, are well-known in the literature for their capabilities to enhance learning with deeper architectures. In this section we report results on one large-scale graph (i.e. OGBN-Arxiv~\cite{wang2020microsoft}) trained on a deeper GNN (i.e. 4 layers), to verify whether \xaidrop scales also to large input graphs. Test accuracy on this dataset has been tested on GCN for all the competitors (apart from Learn2Drop for computational reasons). Results are shown in Table~\ref{tab:ogbn-arxiv}, and confirm the advantage of \xaidrop over its competitors\footnote{\rev{It is important to remind here that our goal is not that of achieving state-of-the-art results using the most recent, complex architectures, for which running competitors would be prohibitively expensive, but showing consistent advantages over alternative solutions when evaluated under the same experimental conditions.}}}

\begin{table}[h!]
    \centering{ 
        \begin{tabular}{l|c|c}
            & \textbf{Model} & \textbf{OGBN-Arxiv} \\
            \toprule
              - & Baseline& \mstd{67.1}{0.8} \\
            \midrule
            \multirow{4}{*}{\rotatebox[origin=c]{90}{Random}} 
                & DropEdge & \mstd{70.5}{1.0} \\
                & \col{DropMess} & \col{\mstd{71.0}{0.6}}  \\
                & DropNode & \mstd{70.7}{0.9} \\
                & \col{DropAggr} & \col{\mstd{69.8}{1.2}} \\ 
            \midrule
            \multirow{1}{*}{\rotatebox[origin=c]{90}{L.}} 
                & BBGDC & \mstd{68.0}{1.1} \\  
            \midrule
            \multirow{3}{*}{\rotatebox[origin=c]{90}{xAI}} 
                & MATE & \mstd{68.8}{1.6} \\     
                & \col{ExPass} & \col{\mstd{70.9}{0.8}} \\
                & ENGAGE & \mstd{71.5}{0.6} \\ 
            \midrule
            \multicolumn{2}{l|}{\col{\xainode}} & \col{\mstd{\textbf{71.7}}{1.2}} \\          
            \bottomrule
        \end{tabular}
    }    
    \caption{\rev{Test set accuracy (in percentage) computed on OGBN-Arxiv trained with GCN across different dropping strategies. The standard deviation is computed on three runs.}}
    \label{tab:ogbn-arxiv}
\end{table}

\section{Extracted explanations}\label{app:extracted_expl}

\begin{figure}[h!]
    \centering
    \includegraphics[width=0.9\linewidth]{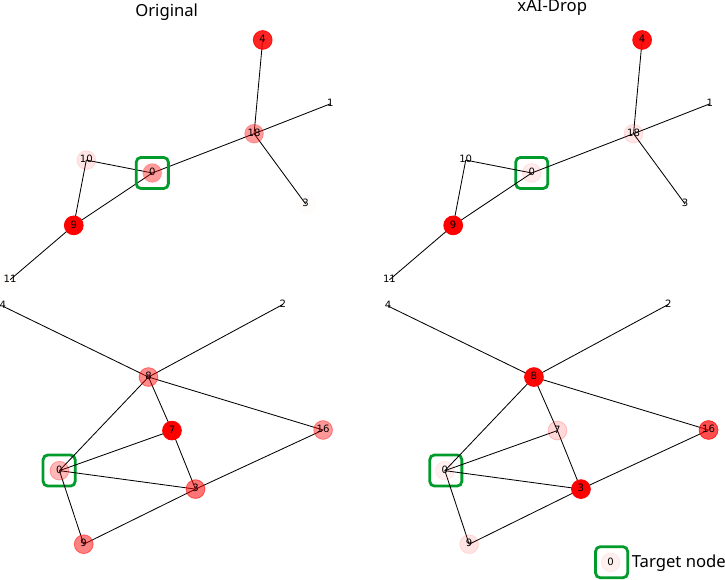}
    \caption{\rev{Examples of explanations generated using a saliency map on a GCN trained on the Cora network.}}
    \label{fig:expl_cora}
\end{figure}

\rev{In Figure \ref{fig:expl_cora}, we present explanations generated using saliency maps on a standard GCN compared to a GCN with our proposed method. The explanations produced after applying our dropping strategies are notably sparser, resulting in clearer visualizations that enable more reliable insights.}

\section{\revsecond{Probability Mapping}}\label{app:exp_mapping_prob}

\revsecond{The mapping of the explanation assessments (i.e. Fidelity $F_{suf}$) is a crucial point in \xaidrop. It enables to convert raw explanation metrics that are usually compressed in a range far from the default dropping probabilities into a range of probabilities $p$ defined in the range $[0,1]$.
We have tested multiple mapping approaches, but two distributions better fill our needs: empirical cumulative distribution and Gaussian distribution (described in Equation \ref{eq:yeo-johnson}). In Table \ref{tab:mapping_prob} we report an empirical comparison across datasets, architectures, and distributions. For the sake of completeness, we report also Uniform distribution which is the one used by random dropping strategies.}
\begin{table}[h!]
    \centering
    \resizebox{\textwidth}{!}{ 
        \begin{tabular}{l@{\hskip 9pt}l@{\hskip 5pt}|c@{\hskip 6pt}c@{\hskip 6pt}c@{\hskip 6pt}|c@{\hskip 6pt}c@{\hskip 6pt}c@{\hskip 6pt}|c@{\hskip 6pt}c@{\hskip 6pt}c}
        &        & \multicolumn{3}{c|}{GCN} & \multicolumn{3}{c|}{GAT} & \multicolumn{3}{c}{GIN}\\
        &Distribution   & Cora & CiteSeer & PubMed & Cora & CiteSeer & PubMed & Cora & CiteSeer & PubMed \\
        \toprule
            &Uniform & 
                \mstd{79.0}{0.3} & \mstd{67.1}{0.5} & \mstd{76.9}{1.2}& 
                \mstd{78.4}{1.2} & \mstd{68.1}{0.7} & \mstd{77.3}{0.7}& 
                \mstd{78.2}{1.0} & \mstd{67.5}{1.0} & \mstd{76.7}{0.8}\\
            &\col{Cumulative} & 
                \col{\mstd{82.6}{0.4}} &
                \col{\mstd{72.6}{0.6}} &
                \col{\mstd{80.9}{0.6}} &
                \col{\mstd{82.5}{0.7}} &
                \col{\mstd{71.7}{0.8}} & 
                \col{\mstd{80.4}{0.7}} &
                \col{\mstd{82.0}{0.7}} &
                \col{\mstd{72.1}{0.6}} & 
                \col{\mstd{79.6}{0.8}}\\
           & \textbf{Gaussian} &
                \mstd{\textbf{82.8}}{0.5} & \mstd{\textbf{74.0}}{0.4} & \mstd{\textbf{81.5}}{0.7} &
                \mstd{\textbf{82.6}}{0.5} & \mstd{\textbf{72.6}}{0.4} & \mstd{\textbf{80.7}}{0.5} &
                \mstd{\textbf{83.0}}{0.4} & \mstd{\textbf{73.0}}{0.6} & \mstd{\textbf{79.6}}{0.7} \\          
        \bottomrule
    \end{tabular}
    }
    \caption{\revsecond{Test accuracy across multiple datasets and architectures tested for node classification task when using different methods for mapping explanation metrics into probability distributions. Note that Uniform refers to random dropping strategies.}}
    \label{tab:mapping_prob}
\end{table}

\end{document}